\pdfoutput=1

\documentclass[11pt]{article}

\usepackage{tikz}
\usetikzlibrary{bayesnet}
\usetikzlibrary{arrows}
\usepackage[]{acl}

\usepackage{times}
\usepackage{latexsym}
\usepackage{algorithm}
\usepackage{algorithmic}
\usepackage{graphicx}
\usepackage{natbib}
\usepackage{amsmath}
\usepackage{tabularx}
\usepackage{amsfonts}
\DeclareMathOperator{\glo}{{global}}
\DeclareMathOperator{\loc}{{local}}
\DeclareMathOperator{\LL}{\mathcal{L}}
\DeclareMathOperator{\kl}{KL}
\DeclareMathOperator{\E}{\mathbb{E}}
\DeclareMathOperator{\doc}{{\mathbf{w}}}
\DeclareMathOperator{\dir}{{Dirichlet}}
\DeclareMathOperator{\multi}{{Multinomial}}

\usepackage[T1]{fontenc}

\usepackage[utf8]{inputenc}

\usepackage{microtype}

\title{Continual Neural Topic Model}

\author{Charu Karakkaparambil James, Waleed Mustafa \\ \textbf{Marius Kloft, Sophie Fellenz} \\
  RPTU Kaiserslautern-Landau \\ Kaiserslautern, Germany \\
  \texttt{surname@cs.uni-kl.de} }

\begin{document}
\maketitle
\begin{abstract}
In continual learning, our aim is to learn a new task without forgetting what was learned previously. In topic models, this translates to learning new topic models without forgetting previously learned topics. Previous work either considered Dynamic Topic Models (DTMs), which learn the evolution of topics based on the entire training corpus at once, or Online Topic Models, which are updated continuously based on new data but do not have long-term memory. To fill this gap, we propose the Continual Neural Topic Model (CoNTM), which continuously learns topic models at subsequent time steps without forgetting what was previously learned. This is achieved using a global prior distribution that is continuously updated. In our experiments, CoNTM consistently outperformed the dynamic topic model in terms of topic quality and predictive perplexity while being able to capture topic changes online. The analysis reveals that CoNTM can learn more diverse topics and better capture temporal changes than existing methods. 
\end{abstract}

\section{Introduction}
Topic models are used to discover the hidden thematic structure in a collection of documents. These models are particularly useful in Natural Language Processing (NLP), supporting a wide range of applications, including information extraction, text clustering, summarization, sentiment analysis, content recommendation, opinion/event mining, and trend analysis \citep{tuan2020capturing, subramani2018novel,nguyen2021enriching, wang2020pandemic,molenaar2024using,wang2011collaborative, avasthi2022topic,churchill2022evolution}. A popular topic model is Latent Dirichlet Allocation (LDA) \citep{blei2003latent}, which represents each document as a collection of topics, with each topic being a distribution over words.

However, document collections are usually not recorded at a singular instance in time. Document collections may span many months or years, during which the topics (and their word distributions) may change. Standard LDA is not designed to handle such dynamic changes.

Dynamic topic models, such as Dynamic LDA \citep{DLDA}, Dynamic Embedded Topic Model  \citep{DETM}, and Dynamic BERtopic \cite{grootendorst_bertopic_2022}, address this issue by capturing the evolution of topics over time. 

Despite their advancements, these models share a limitation: \textit{they require the entire corpus to be available from the start.} 
In real-world applications, new data is generated every day, and it should ideally be processed online in real-time. For example, as data are continuously streamed, there is significant potential in monitoring current topics of interest, detecting emerging trends in social media \citep{sasaki2014online}, analyzing consumer purchase behavior \citep{iwata2009topic}, and tracking urban geo-topics \citep{yao2020tracking}.

To address the need of real-time processing, online topic models have been introduced \citep{alsumait2008line,iwata2010online, zhang2013sparse}, which adapt to data arriving sequentially. 

However, online topic models lack long-term memory and tend to forget previously acquired topic knowledge over time. This is where continual learning becomes relevant. In continual learning, new sub-problems are learned over time without forgetting what was previously learned.

We propose the \textit{Continual Neural Topic Model (CoNTM)}, which uses a global prior distribution to store information over time, while local models capture patterns inherent in the current time step. The proposed model effectively captures global thematic patterns and tracks their temporal evolution over locally defined subsets of the data. The CoNTM maintains high topic quality and low predictive perplexity without losing previously learned information. This is achieved as each time step depends on the global prior, ensuring consistency and coherence over time.

The contributions of the paper are as follows:
 
\begin{itemize}
    \item  We introduce CoNTM, a continual neural topic model to train a sequence of topic models without forgetting what was previously learned.
    
    \item We introduce the CoNTM algorithm, which incrementally updates global topics at each time step $t$, thereby capturing topic dependencies from the previous time step.
     
    \item Through experiments on six diverse datasets, we observe that CoNTM outperforms state-of-the-art DTMs regarding topic quality and predictive perplexity.

    \item Unlike traditional models that require access to all data in advance, we show CoNTM maintains good qualitative performance even when training in a data stream.

\end{itemize}

We discuss related work in Section \ref{sec:related}. The proposed continual modeling methodology is described in more detail in Section \ref{sec:method}. Section \ref{sec:settings} describes the evaluation measures, datasets, and model settings and presents qualitative and quantitative results. Section \ref{sec:conclusion} provides the conclusion.

\section{Related Work}
\label{sec:related}

Tracking the evolution of topics over time has so far been addressed in the research areas of dynamic topic models and online topic models, both of which we review below. We further discuss relevant work in the area of continual learning.

\paragraph{Dynamic Topic Models}


Dynamic topic models assume that the complete corpus is available for training. \citet{rahimi2023antm} differentiates between probabilistic dynamic topic models (PDTMs) and algorithmic dynamic topic models (ADTMs). 

PDTMs are based on generative assumptions. Previous work on PDTMs includes Dynamic LDA \citep{DLDA}, the Dynamic Embedded Topic model (DETM) \citep{DETM}, Dynamic Structured Neural Topic Model with Self-Attention Mechanism \citep{miyamoto-etal-2023-dynamic}, Dynamic Noiseless LDA (DNLDA) \citep{churchill2022dynamic}, modeling discrete dynamic topics \citep{bahrainian2017modeling} and Continuous Time Dynamic Topic Models \citep{wang2012continuous}. DLDA, a probabilistic model, is not based on neural networks and is not scalable to large datasets. DETM combines latent Dirichlet allocation (DLDA) and word embeddings. In DETM, each word is modeled with a categorical distribution parameterized by the inner product of the word embedding and the embedded representation of the topic at each step in time. We compare our model with these baselines in the experiments section. However, unlike our online model, they are batch models and cannot be updated in a continuous data stream.
Another PDTM by \citet{pmlr-v180-tomasi22a} improves rare word inclusion using the correlation-based method and amortized variational inference, making it more efficient for large vocabularies.


ADTMs do not assume a document generation process, but cluster document embeddings and extract topic words using heuristic methods. Unlike PDTMs, which jointly learn topic clusters and embeddings, ADTMs separate these steps. They excel at short texts like Tweets but face challenges with domain-specific corpora due to reliance on pretrained embeddings. Examples include the BERTTopic algorithm \cite{grootendorst_bertopic_2022} using the BERT language model \citep{devlin2018bert}, ANTM \cite{rahimi2023antm}, Dynamite \cite{balepur2023dynamite}, CFDTM \cite{wu-etal-2024-modeling} uses contrastive learning techniques, and Dynamic BERTopic, which extends BERTopic using c-TF-IDF. Other models by \citet{eklund-etal-2022-dynamic, gao2022,boutaleb2024bertrend} also cluster embeddings from pretrained language models (e.g., BERT, GPT) to track topics. 

All these models rely on pretrained embeddings. In contrast, our model learns domain-specific topics even if the above-mentioned pretrained model does not include the target domain.

\paragraph{Online Topic Models}
Online topic models are updated as new data arrives. Previous work includes Online LDA (OLDA) \citep{alsumait2008line}, which incrementally adds new data to the current model rather than requiring access to previous data. The sparse online topic model \citep{zhang2013sparse} uses sparsity-inducing regularization to control the sparsity of latent semantic patterns and employs online algorithms to learn the topical dictionary. The multiscale dynamic topic model \citep{iwata2010online} incrementally updates the model at each epoch using the newly obtained set of documents and the multiscale model from the previous epoch. 
\citet{banerjee2007topic} provide a study on batch and online unsupervised learning. This work is based on statistical topic models and has not yet been adapted to neural network-based topic models with long-term memory capability.

Neural network-based topic models have been studied by \citet{Srivastava2017AutoencodingVI,miao2016neural,burkhardt2019decoupling,dieng2020topic,bianchi2020pre,Srivastava2017AutoencodingVI,reimers2019sentence,grootendorst_bertopic_2022}.

In contrast to our model, these neural topic models do not capture the evolution of topics over time.

\paragraph{Continual Learning}
Continual learning \citep{hadsell2020embracing,zenke2017continual,xu2018reinforced} aims at developing systems that can continuously learn and adapt to new data or tasks over time without forgetting previously learned information. \citet{gupta2020neural} use continual learning for topic models on sparse data, where small collections of documents often lead to incoherent topics. In their work, the authors use multiple-shot task learning with multiple datasets from different domains. 

In contrast, our model applies continual learning to learn new topics within a domain-specific dataset without forgetting previously learned topics.

\section{Methodology}
\label{sec:method}
This section presents the proposed Continual Neural Topic Model (CoNTM).

\subsection{Preliminaries}

The Dirichlet Variational AutoEncoder (DVAE) forms the foundation of the proposed method. In DVAE, a variational autoencoder (VAE) generates the document-topic distribution using an encoder network parameterized by $\theta$. This encoder network effectively captures the thematic structure of the documents by mapping them onto a latent topic space. This means that each document has a document-topic distribution $z\sim \dir(\alpha)$ with a Dirichlet prior. Using a Dirichlet prior is essential as it encourages sparsity in the topic distributions, thereby enhancing the interpretability of the topics. On the other hand, the topic-word distribution, which represents the probability of words given a particular topic, is represented by a decoder network that reconstructs the input documents.

\begin{figure}[ht!]
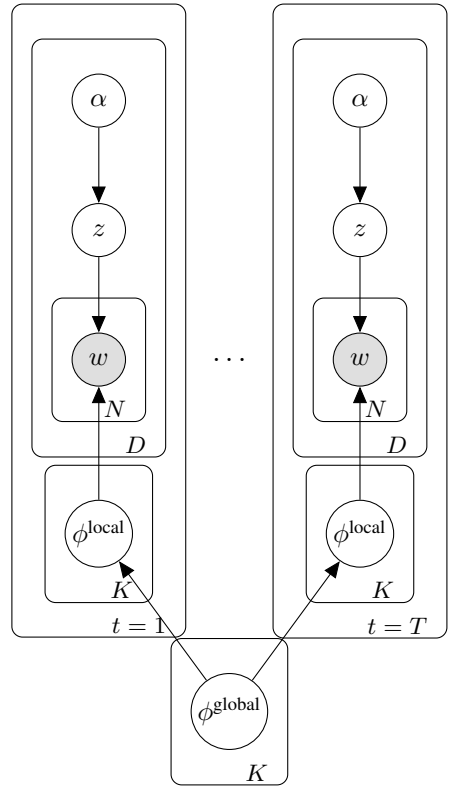

\centering
\tikz{
 \node[obs] (w) {$w$};%
 \node[latent,above=of w] (z) {$z$}; %
 \node[latent,above=of z] (alpha) {$\alpha$}; %
 \node[latent,below=1.5cm of w] (phi) {$\phi^{\text{local}}$}; %
 
 \node[right=1cm of w] (dots){$\hdots$};
 \node[latent,below=4cm of dots] (psi) {$\phi^{\text{global}}$}; %

 \node[obs,right=of dots] (w2) {$w$}; %
 \node[latent,above=of w2] (z2) {$z$}; %
 \node[latent,above=of z2] (alpha2) {$\alpha$}; %
 \node[latent,below=1.5cm of w2] (phi2) {$\phi^{\text{local}}$}; %
 \plate [inner sep=.25cm,yshift=.2cm] {plate1} {(w)} {$N$};
 \plate [inner sep=.25cm,yshift=.2cm] {plate21} {(w2)} {$N$};
 \plate [inner sep=.25cm,yshift=.2cm] {plate2} {(alpha)(w)(z)(plate1)}  {$D$};
 \plate [inner sep=.25cm,yshift=.2cm] {plate22} {(alpha2)(w2)(z2)(plate21)}  {$D$};
 \plate [inner sep=.25cm,yshift=.2cm] {plate3} {(phi)} {$K$};
 \plate [inner sep=.25cm,yshift=.2cm] {plate23} {(phi2)} {$K$};
 \plate [inner sep=.25cm,yshift=.2cm] {plate4} {(psi)} {$K$};%
 \plate [inner sep=.25cm,yshift=.2cm] {plate5} {(plate3)(plate2)} {$t=1$};%
 \plate [inner sep=.25cm,yshift=.2cm] {plate25} {(plate23)(plate22)} {$t=T$};%
 \edge {z} {w} ;
 \edge {alpha} {z};
 \edge {phi} {w};
 \edge {psi} {phi};
 
 \edge {z2} {w2} ;
 \edge {alpha2} {z2};
 \edge {phi2} {w2};
 \edge {psi} {phi2};
 }
\caption {The Graphical Model of our approach. The local models of each time slice $t$ are connected by the global parameters $\phi^{\text{global}}$, which captures topic dependencies from the previous time step.}
\label{fig:gm}
\vspace{-0.5cm}
\end{figure}

\subsection{Continual Neural Topic Model (CoNTM)}
In CoNTM, we model documents as arriving in continuous time slices, with each slice characterized by slightly varying topics. These topics are interconnected through a global topic set, allowing for minor temporal adjustments to the global topics at each time step $t$. This approach implicitly captures topic dependencies from the previous time step via the global distribution while offering the advantages of a reduced number of parameters and increased model flexibility as compared to approaches that explicitly model each transition. 

Formally, a document at time $t$ is modeled as a mixture of local topics $\phi_t^{\loc} = (\phi_{t,1}^{\loc}, \ldots, \phi_{t,K}^{\loc})$, where each topic $\phi_{t,k}^{\loc}$ is a probability distribution over the vocabulary. We further assume that the local topics are derived from a set of global topics $\phi^{\glo}$ such that $\phi_t^{\loc} = g(\phi^{\glo}, \Delta \phi_t^{\loc})$, where $g$ is a transformation function that applies a perturbation $\Delta \phi_t^{\loc}$ to the global topics to obtain the local topics. This assumption ensures that local topics remain consistent over time. The graphical model is shown in Figure \ref{fig:gm}. 

\subsubsection{Generative Process}
The algorithm \ref{alg:algorithm1} outlines the assumed generative process of our data. According to this generative process, the marginal distribution of a document $\doc$ at time $t$ is given by 
\begin{align*} p(&\doc | \alpha, \phi^{\glo},\Delta \phi_t^{\loc}) = \\
 &\int_z p(z \mid \alpha) \prod_{n=1}^{N} p(\doc_n \mid z, g(\phi^{\glo}, \Delta \phi_t^{\loc})) \, dz
\end{align*}

\begin{algorithm}[t!]
\caption{data generative algorithm}
\label{alg:algorithm1}
\begin{flushleft}
\textbf{Input}: Number of time slices $T$, documents $\doc$ \newline
\textbf{Output}: Local Topics $\{\phi_t^{\loc}\}_{t=1}^{T}$
\end{flushleft}
\begin{algorithmic}[1]
\FOR{$t = 1 \hdots T$}
    \STATE $\phi_{t}^{\loc} \leftarrow g(\phi^{\glo}, \Delta \phi_t^{\loc})$  
    \FOR{ each document $\doc$}
        \STATE Draw a document topic distribution:
        \vspace{-1em}
        \begin{equation*}
            z \sim \dir(\alpha)
        \end{equation*}
        \vspace{-1em}
        \FOR{ each word index $n$}
            \STATE Draw a word:
            \vspace{-1em}
            \begin{equation*}
                \doc_n \sim \multi(1,\phi^{\loc}_tz)
            \end{equation*}
            \vspace{-1em}
        \ENDFOR
    \ENDFOR
\ENDFOR
\end{algorithmic}
\end{algorithm}

where \(p(z \mid \alpha)\) is the Dirichlet distribution over topic proportions, and 
$p\left(\mathbf{w}_n \mid z, g\left(\phi^{\text {global }}, \Delta \phi_t^{\text {local }}\right)\right)$ is the multinomial distribution over words given the local topics. 

\subsubsection{Variational Inference}
The posterior inference over the parameters $z$ is intractable \citep{Srivastava2017AutoencodingVI}. We thus resort to the DVAE framework \citep{burkhardt2019decoupling}. That is, we assume a variational distribution $q_{\theta}(z)$ on the random variable $z$ parameterized by a free parameter $\theta$, which is learned by maximizing the Evidence Lower Bound (ELBO)  
\begin{align*}
\LL(\theta,\doc| \phi^{\glo}, \Delta &\phi^{\loc}_t) =\\
    -\kl(q_\theta(z) &\,\|\,p(z\mid\alpha))\, + \\
    \E_{q_{\theta}(z)}&[\log p(\doc \mid z, \phi^{\glo}, \Delta\phi^{\loc}_t)].
\end{align*}
The form of $ q_{\theta}(z)$ is characterized by an encoder network $\alpha_{\theta}(\doc)$ parameterized by $\theta$. Specifically, the variational distribution is defined as $\dir(\alpha_\theta(\doc))$.
\begin{algorithm}[t!]
\caption{CoNTM algorithm}
\label{alg:algorithm}
\begin{flushleft}
\textbf{Input}:  Number of time slices $T$, stream of documents $\{\doc^{i,1}\}_{i=1}^{n_1}, \cdots, \{\doc^{i,T}\}_{i=1}^{n_T}$ , number of Topics $K$, and number of training steps $J$\\
\textbf{Output}: Topics: $\{\hat{\phi}_t^{\loc}\}_{t=1}^T$, $\hat{\phi}^{\glo}$ \\
\end{flushleft}
\begin{algorithmic}[1] 
\FOR{$t = 1 \hdots T$}
\STATE Initialize $\theta$
\STATE Initialize $\Delta \hat{\phi}^{\loc}_t$
\STATE $\{\doc^{i,t}\}_{i=1}^{n_t}\leftarrow $  documents arrived at time $t$
\FOR{$j = 1 \hdots J$}
\STATE Update $\theta$ by the gradient 
\vspace{-1em}
\[\nabla_{\theta}\sum_{i=1}^{n_t} \LL(\theta,\doc^{i,t} \mid \hat{\phi}^{\glo}, \Delta \hat{\phi}^{\loc}_t) \] \vspace{-1em}
\STATE Update $\Delta \hat{\phi}^{\loc}_t$ by the gradient \vspace{-1em}
\[\nabla_{\Delta \hat{\phi}^{\loc}_t}\sum_{i=1}^{n_t} \LL(\theta,\doc^{i,t} \mid \hat{\phi}^{\glo}, \Delta \hat{\phi}^{\loc}_t) \] \vspace{-1em}
\ENDFOR
\STATE Set $\hat{\phi}^{\loc}_t \leftarrow \hat\phi^{\glo} + \Delta \hat{\phi}^{\loc}_t$ 
\STATE Set $\hat\phi^{\glo} \leftarrow(1-\rho_{t}) \hat\phi^{\glo}+\rho_{t} \hat\phi^{\loc}_t $ 
\ENDFOR
\end{algorithmic}
\end{algorithm}


Optimizing the ELBO with respect to $\phi^{\loc}_t$ and $\phi^{\glo}$ can be challenging due to the simplex constraint ($\sum_{w}\phi_{k,w} = 1$). Therefore, we follow the DVAE topic model approach by introducing unconstrained variables $\Delta\hat{\phi}^{\loc}_t$ and $\hat{\phi}^{\glo}$. Using these variables, we define the probability of the \(n\)-th word \(\doc_n\) in a document as follows:
\begin{align*}
    p(\doc_n=w \mid  z&, \hat{\phi}^{\glo}, \Delta\hat{\phi}^{\loc}_t)=\\
    &[\sigma( g(\hat{\phi}^{\glo}, \Delta\hat{\phi}^{\loc}_t)\cdot z) ]_{w},
\end{align*}
where $\sigma$ denotes the softmax function. Notably, the softmax normalization is performed after mixing the resulting local topics with the topic weights \(z\). This approach is shown to enhance the model's expressive power by treating the word distribution as a product of experts ~\citep{Srivastava2017AutoencodingVI}. 

\subsubsection{Global Topic Parameters}

We now turn our attention to the function \( g \), which links the local to the global topic parameters. While our approach accommodates a general form for \( g \), we assume a simple form:
\[ \hat{\phi}^{\loc}_t = g(\hat{\phi}^{\glo}, \Delta\hat{\phi}^{\loc}_t) := \hat{\phi}^{\glo} + \Delta\hat{\phi}^{\loc}_t. \]
We further introduce the constraint $\sum_{t=1}^T \Delta\hat{\phi}^{\loc}_t = 0$. This can be interpreted as $\Delta\hat{\phi}^{\loc}_t$ being realized from a centered probability distribution. The choice of $g$ and the constraint on $\Delta\hat{\phi}^{\loc}_t$ has the advantage that
\[ \hat{\phi}^{\glo} = \frac{1}{T} \sum_{t=1}^T \hat{\phi}^{\loc}_t. \]
Thus, inference is simplified by only computing the local topics $\hat{\phi}^{\loc}_t$ and obtaining the global topics by taking their average. Since we are interested in continual learning, we replace taking the average with a running average of the local topics, that is 
\[
\hat{\phi}^{\glo} = \frac{t-1}{t}\hat{\phi}^{\glo} + \frac{1}{t}\hat{\phi}^{\loc}_t.
\]
To control the amount of updates to $\hat{\phi}^{\glo}$ over time, we further replace $\frac{t-1}{t}$ (resp. $\frac{1}{t}$) by $1-\rho_t$ (resp. $\rho_t$) for $\rho_t \in (0,1)$. For instance, we can set \(\rho_t = \frac{1}{(\tau_0 + t)^{\kappa}}\), where \(\kappa \in (0.5, 1]\) and \(\tau_0 \ge 0\) as in \citet{hoffman2010online}. Here, \(\kappa\) controls the rate of forgetting the old estimate of the global topics, while \(\tau_0\) slows down the updates in the early steps. Selecting proper values for \(\tau_0\) and \(\kappa\) ensures that the influence of local updates decreases over time, allowing the model to converge to a global topic-word distribution. Algorithm~\ref{alg:algorithm} summarizes the learning procedure.

\section{Experiments}
\label{sec:settings}
This section compares the proposed model to four state-of-the-art models. We present both quantitative results and a detailed qualitative analysis of how topics evolved over time.

\subsection{Evaluation of Dynamic Topic Models}
We use topic coherence (TC) and topic diversity (TD) to evaluate the dynamic topic model. A commonly used TC metric is the Normalized Point-wise Mutual Information (NPMI) \citep{bouma2009normalized,lau2014machine, DETM, dieng-etal-2020-topic,bianchi2020cross}. A temporal reference corpus was utilized to determine the NPMI score for the topics. This means that the NPMI score for a topic at a specific timestamp is calculated using the reference corpus available up to that point in time. \citet{burkhardt2019decoupling} proposed Topic Redundancy (TR), a measure that calculates the average occurrences of a top word in other topics. The topic diversity is calculated as $TD = 1-TR$. The redundancy for topic k is given below:
$$
TR(k)=\frac{1}{K-1} \sum_{i=1}^N \sum_{j \neq k} P\left(w_{i k}, j\right).
$$
Here, $P\left(w_{i k}, j\right)$ equals one if the $i$th word of topic $k, w_{i k}$, occurs in topic $j$ and otherwise zero. $K-1$ is the number of topics excluding the current topic.  

To ensure that the topic quality is not affected by the occurrence of too few or too many topics, it is normalized based on the total number of topics in each timestamp \cite{rahimi2023antm}: 
$$
\text { TQ }=\frac{1}{k} \sum_{i=0}^{k-1} \mathrm{TC}_i \times \mathrm{TD}_i \times \frac{T_i}{\mathrm{~T}_i^{\text {max }}}.
$$
Here $\mathrm{TC}_i$ and $\mathrm{TD}_i$ represent topic coherence and diversity in timestamp $i$. Additionally, $T_i$ represents the number of topics within timestamp $i$, and $\mathrm{~T}_i^{\text {max }}$ indicates the highest number of topics observed across all years. To track topic changes over time, we use the recently proposed Temporal Topic Smoothness (TTS) measure \citep{karakkaparambil-james-etal-2024-evaluating}. The TTS indicates whether the topic transition is abrupt or gradual. 

Finally, we also use predictive perplexity (PPL) on unseen future timestamps \citep{wang2012continuous}, a standard evaluation measure for assessing the performance of probabilistic language models. It quantifies the model's ability to predict topics in future timestamps by calculating the average log-likelihood of a sequence and then exponentiating its negative value. Lower perplexity values indicate better predictive model performance.

\subsection{Datasets}
Our study is conducted on six widely recognized datasets within the field (see Table \ref{tab:Corpus_Statistics} in Appendix \ref{sec:appendixB}). The first is a collection of articles from the New York Times \citep{AB2/GZC6PL_2008}, covering a span of two decades, specifically from 1987 to 2007. The second, the UN corpus \cite{DVN_0TJX8Y_2017}, encompasses a temporal range of five decades, extending from 1970 to 2020, and comprises statements from the general debates over this period. The third dataset under study is the NIPS corpus \cite{swami_2020}, which includes the entirety of the NIPS conference published between 1987 and 2019. The fourth dataset contains around 14,000 tweets from @NASA's Twitter account, spanning a period of over four years, from 2018 to 2022. The fifth dataset includes a collection of 16,000 documents, consisting of arXiv titles and abstracts \citep{arxiv_org_submitters_2023}, covering the years 2012 to 2024. The final dataset consists of the DBLP archive \citep{ley2002dblp}, which includes 168,000 scientific articles (titles and abstracts) published between 2000 and 2020. Additionally, the details of the preprocessing steps can be found in Appendix \ref{sec:appendixB}.

\setlength{\tabcolsep}{3pt}
\begin{table*}[ht!]
\begin{center} 
\begin{tabular}{l|ccc|ccc|ccc|ccc|ccc|}  
 & \multicolumn{15}{c}{\textit{50 topics}}  \\\cline{2-16}
 & \multicolumn{3}{c}{\textbf{CoNTM (ours)}} & \multicolumn{3}{|c|}{\textbf{DETM}} & \multicolumn{3}{c}{\textbf{DLDA}} & \multicolumn{3}{|c|}{\textbf{DBERTopic}} & \multicolumn{3}{|c|}{\textbf{DNLDA}}\\\hline
\textbf{Dataset} & TC & TD & TQ & TC & TD & TQ  & TC & TD & TQ &  TC & TD & TQ & TC & TD & TQ  \\ 
\hline 
NIPS&.087 &.969 &.084 & -.009 & .970 &-.009 &.097 & .980 &.095 & .058 & .343 & .032& .023&.997 &.022 \\
NYT& .174&.991 & .173&  .137 & .987 &.135 &.122 & .939 & .116 &.117  &.797  &.082 &.069 & .997& .069\\

UN & .085&.863 & .074& -.045 & .958 &-.043 &.096 & .942 &.091 & .057 &.515  &.024 &.034 & .994&.034 \\

Tweets &-.003& .907& -.006& -.008 & .982 &-.008 & .036& .965 &.019 &.114  &.912  &.096 &-.135 &.994 &-.135\\

Arxiv & .102& .974& .099& .069 & .963 &.067 &.084 & .966 & .082& .064 & .962 & .059& -.004& .996&-.004\\

DBLP & .118&.969 & .115& .073 & .966 & .071&.101 & .958 &.097 &.093  &.951  &.086 & .041& .998&.040\\

\hline
Av. Rank &\textbf{1.6}  &3.1  &\textbf{1.6} & 3.5 & 2.6 &3.5 &1.8 & 3.5 &1.8 &3 &4.6  & 3.1 & 4.6&\textbf{1} & 4.5\\
\hline
\end{tabular} 

\end{center}
\caption{The table compares the average performance over three runs of five topic modeling algorithms: CoNTM, DETM, DLDA, Dynamic Bertopic, and DNLDA, based on topic coherence (TC), topic diversity (TD), and topic quality (TQ). These evaluations were conducted on six diverse datasets. The CoNTM model maintains good topic quality as a continual learning model, provided it has a sufficiently large dataset.}
\label{tq}
\vspace{-0.2cm}
\end{table*}

\subsection{Models}
The CoNTM model is evaluated alongside four distinct baseline models: Dynamic Embedded Topic Model (DETM), Dynamic Latent Dirichlet Allocation (DLDA), Dynamic BERTopic (DBERTopic), and Dynamic Noiseless LDA (DNLDA).  DLDA and DNLDA offer more traditional probabilistic approaches with a focus on topic evolution, while DETM and Dynamic BERTopic leverage embeddings to capture semantic changes. For the CoNTM, a learning rate of 0.01 is adopted. The model employs the Adam optimizer and divides the dataset into an 80\% training set, a 10\% validation set, and a 10\% test set (see Appendix \ref{sec:appendixA} for the hyperparameter settings of other models). For the experiment, a $\kappa$ value of 0.7 and a $\tau$ value of 1 were used, and sensitivity analysis is presented in Appendix \ref{Sensitivity Analysis of Rho}. The number of topics used across all models in this study is uniformly set to 50. Experimental results for the models configured with 20 topics are shown in Appendix \ref{Stability Across Varying Topic Size}, which shows the stability across varying topic sizes. Additional experiments on generative predictive perplexity are presented in Appendix \ref{PPL}.

\subsection{Quantitative Results}
This section compares the CoNTM model to four different baselines, focusing on coherence, diversity, smoothness, and temporal aspects while highlighting model strengths on domain-specific datasets.

\begin{figure*}[h!]
\includegraphics[scale=0.31]{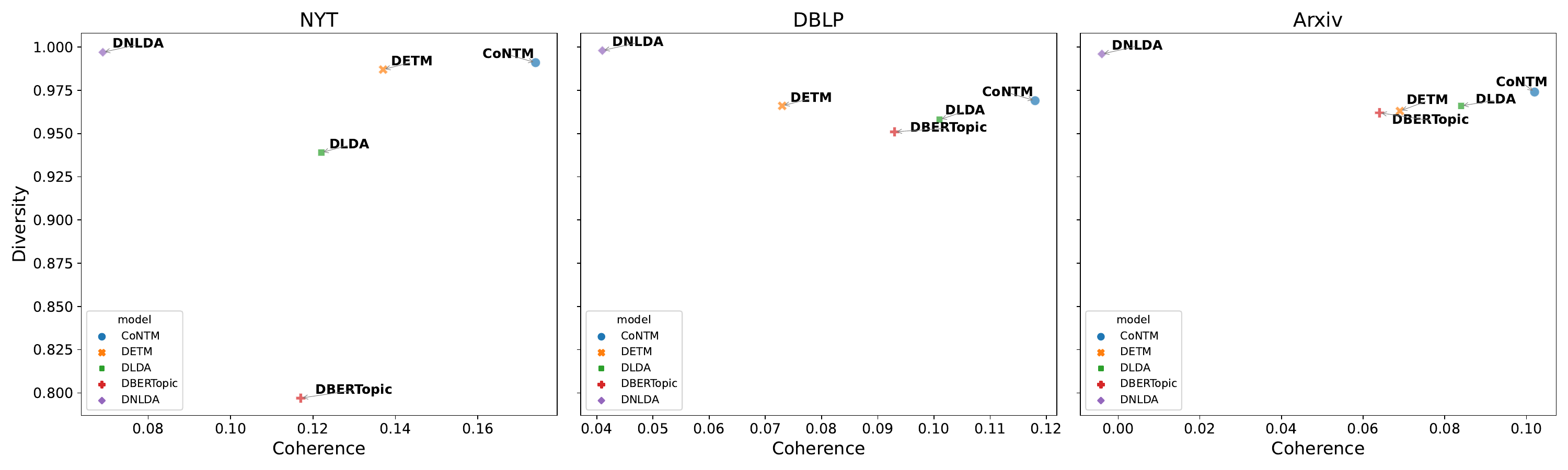} 
\caption{The figure shows the model performance quantitatively for the NYT, BDLP, and Arxiv datasets. The top-right corner indicates that the model achieves high topic quality and low predictive perplexity. Our model (CoNTM) outperformed the other models for these datasets in terms of both coherence and diversity.}\label{fig:all_coherence_vs_diversity}
\vspace{-0.4cm}
\end{figure*}

\begin{figure*}[h!]
\includegraphics[scale=0.45]{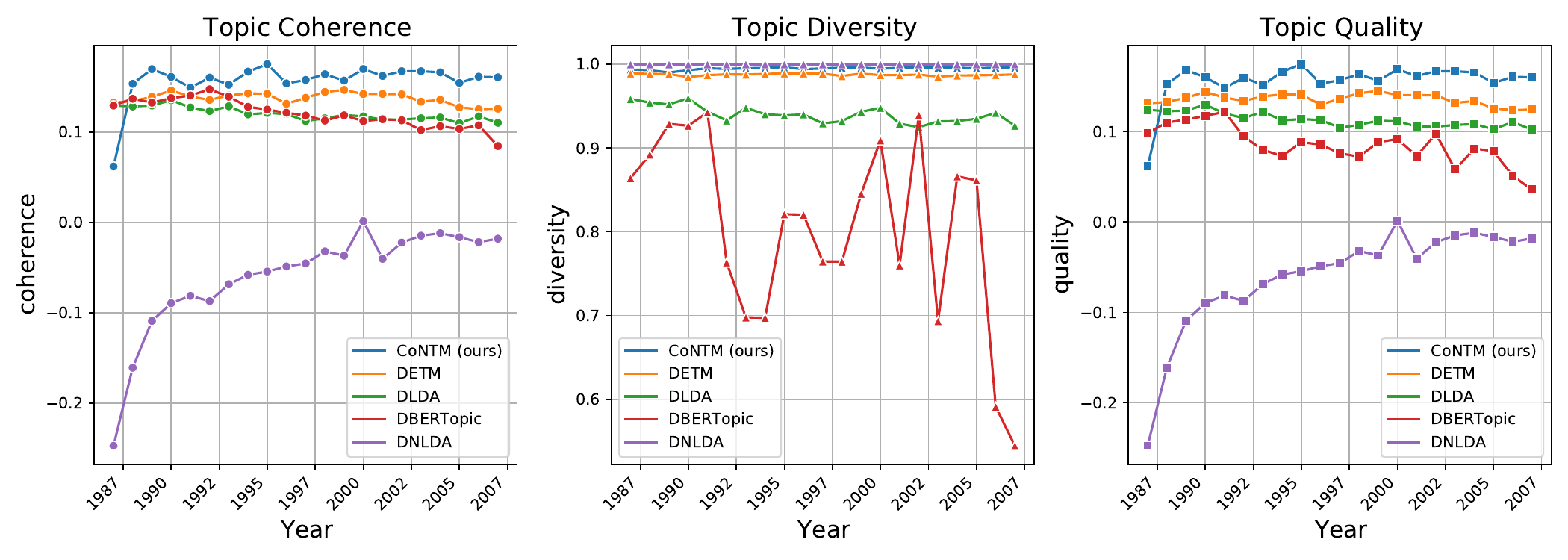} 
\caption{The figure illustrates how coherence, diversity, and topic quality change over the years for the NYT Dataset. Our model shows increasing topic coherence over the year and outperforms other models. Both the DETM and DLDA show stable coherence, while the Dynamic BERT model shows a decrease in coherence over the year. }\label{fig:NYT_tq}
\vspace{-0.4cm}
\end{figure*}

\textbf{Topic Coherence vs Diversity:} After analyzing topic coherence, diversity, and quality, we found that our CoNTM model exhibits good topic quality with an average rank of 1.6 among the five models (see Table \ref{tq}) and maintains good topic diversity score (see Figure \ref{fig:all_coherence_vs_diversity}) on large datasets. 
The value presented in the table is the average of three random seed values. The average rank for each model is computed by averaging its ranks across all datasets. CoNTM, DLDA, and DBERTopic showed moderate topic coherence, whereas DETM and DNLDA presented varied results (see Table \ref{tq}). Additionally, as shown in Figure \ref{fig:all_coherence_vs_diversity}, the CoNTM model achieved good topic quality and good topic coherence on NYT, DBLP, and Arxiv datasets, which had sufficiently large documents. The alternate datasets, NIPS, UN, and Tweets, are described in Appendix \ref{Topic Coherence vs Diversity}. Models in the top-right corner have good topic quality and good topic diversity, indicating good performance.

For smaller datasets, such as Tweets, which consist of approximately 9,703 documents, the Dynamic BERTopic model exhibits good performance based on topic quality. This model uses pre-trained word embeddings, significantly improving the topic quality compared to other models in small datasets. However, it is observed that as the size of the dataset increases (see Table \ref{tab:Corpus_Statistics}), the performance of the Dynamic BERTopic model tends to decline, except for Tweets. 
Furthermore, for the NIPS dataset, the CoNTM model exhibits moderate topic quality compared to other models. This is due to the insufficient documents available at the early timestamp for the NIPS dataset. This affects topic quality in subsequent timestamps, as our model is continual. 

We can also observe that for Dynamic BERTopic, the topic quality is lower on domain-specific datasets like the UN compared to general datasets. This may be because large language models are not trained on domain-specific datasets, leading to a decline in topic quality when using pre-trained word embeddings.

In summary, the CoNTM model consistently demonstrates strong topic quality, highlighting its effectiveness in extracting coherent and meaningful topics from the data. Notably, this performance is achieved without the assumption that all data are available from the start. Instead, CoNTM is capable of handling scenarios where data arrive incrementally.

\paragraph{Temporal Quality and Smoothness}

An analysis of the NYT dataset from 1987 to 2007 was conducted to determine the temporal quality of the dataset. Each model's coherence, diversity, and overall quality were tracked over the years (see Figure \ref{fig:NYT_tq}). Our model (CoNTM) maintained stable performance across the years, with an increase in coherence and quality towards the later years. As a result of incorporating continual learning into the model, the quality of topics has increased over the years, in contrast to other models. DBERTopic and DNLDA exhibit fluctuations in performance. While topic coherence for our model increases with time, it decreases for Dynamic BERTopic. 

\begin{figure*}[t!]
\includegraphics[scale=0.48]{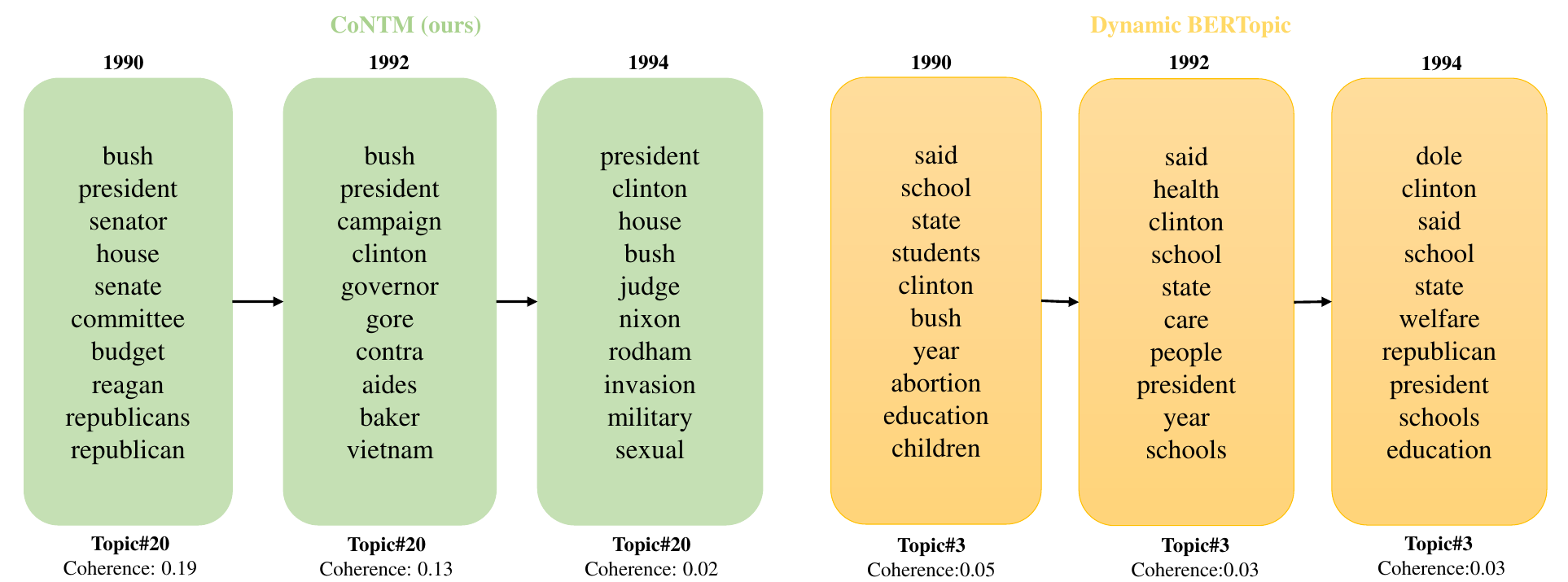} 
\caption{The figure shows the progression of two topics from CoNTM and Dynamic BERTopic in the NYT dataset from 1990 to 1994. Our focus is on the US election in 1992, where \textit{Clinton} was elected as president. CoNTM demonstrates better topic coherence, capturing \textit{Clinton} as a top word in 1992, while Dynamic BERTopic shows little change over time.}\label{fig:nyt_DNTM_DBERTopics_topics}
\vspace{-0.6cm}
\end{figure*}

Additionally, Dynamic BERTopic shows a drastic change in topic diversity, which negatively impacts topic quality. The DETM and DLDA show stable diversity and quality over time. Additionally, the DNLDA model shows an improvement in topic quality, though it remains significantly lower than the others. Furthermore, the temporal quality of UN from 1970 to 2020 can be seen in Figure \ref{fig:un_tq}. The temporal characteristics of the Arxiv dataset are illustrated in Figure \ref{fig:arxiv_tq}. Similarly, the temporal characteristics of the NIPS dataset are shown in Figure \ref{fig:nips_tq} in Appendix \ref{sec: Qualitative Assessment on UNDebates Dataset}.

\setlength{\tabcolsep}{1pt}
\renewcommand{\arraystretch}{1.2} 
\begin{table}[ht!]
\centering
\begin{tabular}{l|cccccc}
\hline
\textbf{Models} & \textbf{NIPS} & \textbf{NYT} & \textbf{UN} & \textbf{Tweets} & \textbf{Arxiv}  & \textbf{DBLP}\\
\hline
CoNTM  & .368 & \textbf{.569}& \textbf{.496} & .191 & \textbf{.498} & \textbf{.599}\\
DETM  & .802 &.646 &.870  &.820  &.864  & .866 \\
DLDA  &.602  & .713& .635 & .372 & .655 & .748\\
DBERTopic  & .272 & .360& .307 & .186 & \textbf{.495} & \textbf{.547}\\
DNLDA  & .016 & .008& .019 & .017 & .013 & .015\\
\hline
\end{tabular}
\caption{The tables display Temporal Topic Smoothness (TTS) scores for four models across various datasets. Our model averages a TTS score of 0.51, except for the Tweets dataset, indicating that its topic transitions are balanced. The bolded number indicates the TTS score closest to $0.50 \pm 10$.}
\label{table:tts}
\vspace{-0.6cm}
\end{table}

For each dataset, the score for temporal topic smoothness (TTS) can be found in Table \ref{table:tts}. On average, the CoNTM model has a TTS score of 0.49, with an exception on the Tweets dataset. The Tweets have a low TTS score because the dataset lacks sufficient documents to learn more coherent and diverse topics. In summary, while the topics change gradually, the transitions are not completely smooth, allowing us to observe the evolution of topics. This is because our model learns new topics at each timestamp without forgetting previously learned information.

\subsection{Qualitative Results}
\label{Qualitative Results}
This section qualitatively analyzes the "politics" topic as it evolves over time. Additional analysis is shown in Appendices \ref{sec: Additional Qualitative Assessment on NYT Dataset} and \ref{sec: Qualitative Assessment on UNDebates Dataset}, indicating that CoNTM and DETM produce more coherent topics for the NYT dataset while tracking topic evolution. 

\textbf{Evolving Topics:} CoNTM effectively captures evolving topics,  as shown in Figure \ref{fig:nyt_DNTM_DBERTopics_topics}, highlighting the 1992 election of \textit{Clinton}. Key topic words during this timestamp include "bush", "president", "campaign", and "clinton". In contrast, Dynamic BERTopics shows little change over time, not adequately representing the events per year. Figure \ref{fig:NYT_emergingTopic} in Appendix \ref{Emerging Topics} additionally illustrates the evolving topic "politics" with the word probability on the y-axis and timestamp on the x-axis. We observe an increase in the word probability of \textit{Clinton} in 1997, which corresponds to his inauguration for a second term as the 42nd President of the United States. His final years in office were from 1999 to 2000. See Appendix \ref{Emerging Topics} for an additional example from the UN dataset.

\textbf{Topic Coherence:} The topics generated by CoNTM exhibit high coherence (Figure \ref{fig:nyt_DNTM_DBERTopics_topics}), with clear connections among the words. For instance, the transition to \textit{Clinton's} 1992 election highlights political themes. This suggests that CoNTM effectively captures topic changes over time, maintaining a strong thematic connection. While Dynamic BERTopics can capture a broad range of topics, the coherence within each topic is lower than CoNTM. In conclusion, CoNTM excels in detecting emerging topics with good topic coherence. 

\section{Conclusion}
\label{sec:conclusion}
We have presented a novel Continual Neural Topic Model (CoNTM), a DVAE-based method for topic modeling that continuously learns evolving topics without forgetting previously learned information. We evaluate CoNTM using datasets from various domains, including news, politics, science, and Tweets, through both quantitative and qualitative analysis. Furthermore, CoNTM demonstrates the ability to track temporal evolution in real-time sequential data. Notably, CoNTM outperforms the DTM models when dealing with large datasets, even though the CoNTM model does not assume all data to be available from the start. In the future, we want to extend our model to handle datasets of varying sizes by incorporating word embeddings.

\section*{Limitations}
The Continual Neural Topic Model, while offering significant advances in topic modeling, especially in its ability to capture the evolution of a topic over time, comes with some limitations. First, these models often require a large amount of data to train effectively. In scenarios where data is sparse or where topics evolve rapidly, the model might struggle to learn meaningful patterns. Second, while these models are designed to capture topic evolution over time, they may not always accurately reflect rapid shifts in topics, particularly in fast-changing domains like social media or news. Third, evaluating the quality of topics and their temporal evolution remains a challenge, as traditional topic coherence metrics may not fully capture the semantic shifts over time, making reliable assessment and comparison with static models difficult.
\bibliography{anthology,custom}
\bibliographystyle{acl_natbib}

\appendix

\section{Model Settings}
\label{sec:appendixA}

This section describes the configuration of hyperparameters for each topic model under study. For the DETM (Dynamic Embedded Topic Model), word representations are derived using a skip-gram model with a 300-dimensional vector space. The DETM utilizes the perplexity score on the validation set as a criterion for termination. The learning rate for DETM is set at 0.001, and the hyperparameters delta, sigma, and gamma are fixed at 0.005, as recommended by the original authors. A uniform batch size of 100 is applied across all datasets. Additionally, the document corpus is segmented into training (80\%), validation (10\%), and testing (10\%) subsets.

In the process of DLDA model training, the Gensim Python library's wrapper for Dynamic Topic Models (DTM) is utilized. Using this approach, all datasets are partitioned annually, thereby encapsulating each year's documents within a single temporal slice. For every dataset, the model undergoes 50 iterations, employing an alpha parameter set to 0.01. This alpha value is a critical hyperparameter in the Latent Dirichlet Allocation (LDA) models, influencing the degree of sparsity in the document-topic distributions across each time slice. Additionally, the top\_chain\_var parameter is set to 10. This setting plays a pivotal role in determining the variability in topic evolution over time within the DTM framework.

The default setting for Dynamic BERTopic was used, and the parameter evolution\_tuning was set to true to display the topic evolution. Similarly, the default DNLDA model settings were used, tnd\_iterations and lda\_iterations being 500. The lda\_beta value is 0.01, and the topic\_depth value is 100. The results shown for all models are the average performance over three runs.

\section{Preprocessing Details}
\label{sec:appendixB}
To preprocess the datasets, the text was converted to lowercase, and both stopwords and punctuation were removed. As part of tokenization, we used Spacy \cite{spacy2}. For removing words that are present in fewer than or more than min/max percent of documents, we used Scikit-learn \citep{kramer2016scikit}. The UN and NIPS corpus has been broken down into paragraphs, with each paragraph being treated as a separate document. The statistical analysis of the document corpus is shown in table \ref{tab:Corpus_Statistics}.

\setlength{\tabcolsep}{2pt}
\begin{table*}[hbt]
\begin{center} 
\begin{tabular}{l|rrrrrr}
\textbf{Dataset}  & \multicolumn{1}{c}{\textbf{    NYT}}  & \multicolumn{1}{c}{\textbf{UN}} & \multicolumn{1}{c}{\textbf{NIPS}} & \multicolumn{1}{c}{\textbf{NASA}}& \multicolumn{1}{c}{\textbf{Arxiv}}& \multicolumn{1}{c}{\textbf{DBLP}}\\ 
\hline
Domain & \multicolumn{1}{c}{News}  & \multicolumn{1}{c}{Politics} & \multicolumn{1}{c}{Science}  & \multicolumn{1}{c}{Twitter}  & \multicolumn{1}{c}{Science} & \multicolumn{1}{c}{Science}\\
Number of Docs  & {273,938}  & {273,398} & {276,657} & {9,703}& {129,984} & {151,232}\\
Vocab Size  & 9,046 & 6,365 & 6,278 & 4,290 & 3,322&  3,162\\ 
Timestamp & 21 & 51 & 11 & 5 & 13 & 11\\%
 min\_df  & 0.3\% & 0.05\% & 0.05\% &  0.05\% &  0.3\% & 0.3\%\\
  max\_df  & 95\% & 95\% & 95\%  &  95\% & 95\% &  95\% \\
\end{tabular} 
\end{center}
\caption{Statistical analysis of corpus data. It demonstrates the datasets that vary in domain, number of documents, and vocabulary size. The NIPS is from \cite{swami_2020}, NYT is from \cite{AB2/GZC6PL_2008}, UN is from ~\cite{DVN/0TJX8Y_2017}, Arxiv is from \cite{arxiv_org_submitters_2023}. Additionally, it displays the vocabulary size of documents that contain less than min\_df percent of words and more than max\_df percent of words.}
\label{tab:Corpus_Statistics}
\end{table*}

\begin{figure*}[ht!]
\includegraphics[scale=0.49]{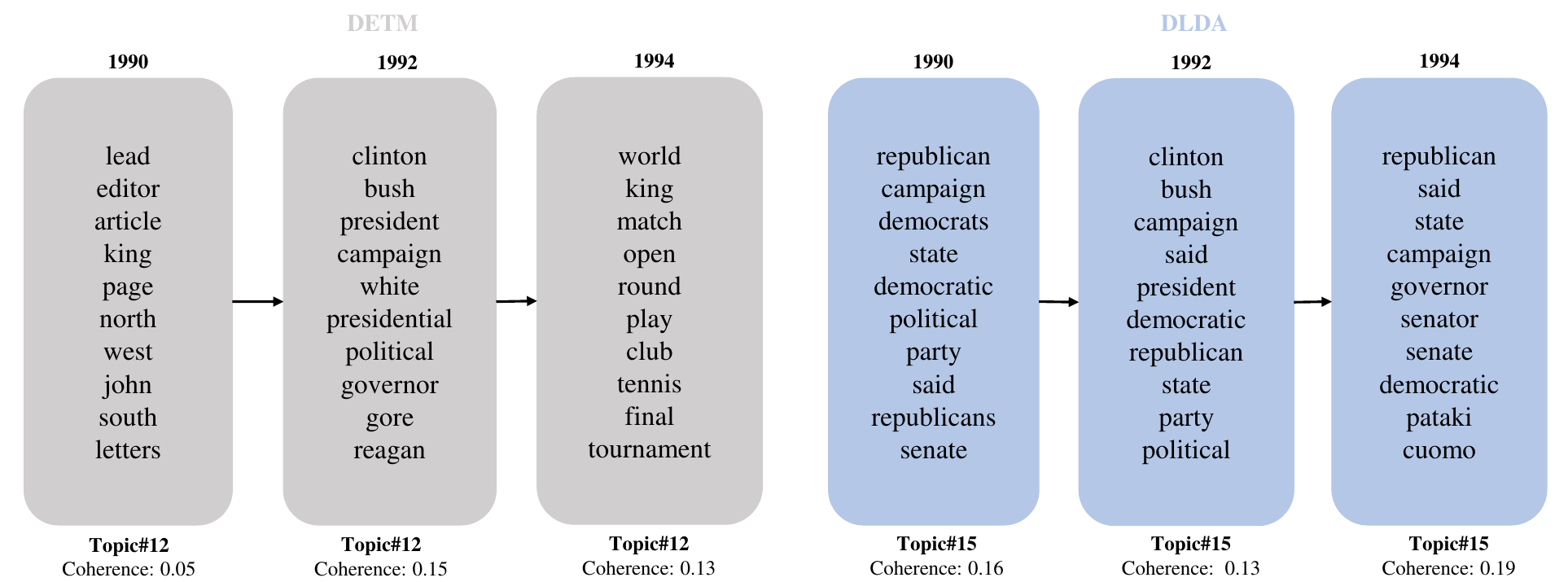} 
\caption{The figure illustrates the progression of topics\#12 (DETM) and topic\#15 (DLDA) in the NYT dataset, spanning the years 1990 to 1994, both of which focus on politics (election of Clinton in 1992). The DLDA models reflect stable and clear thematic focus topics.}\label{fig:nyt_DETM_DLDA_topics}
\end{figure*}

\begin{figure*}[ht!]
\includegraphics[scale=0.49]{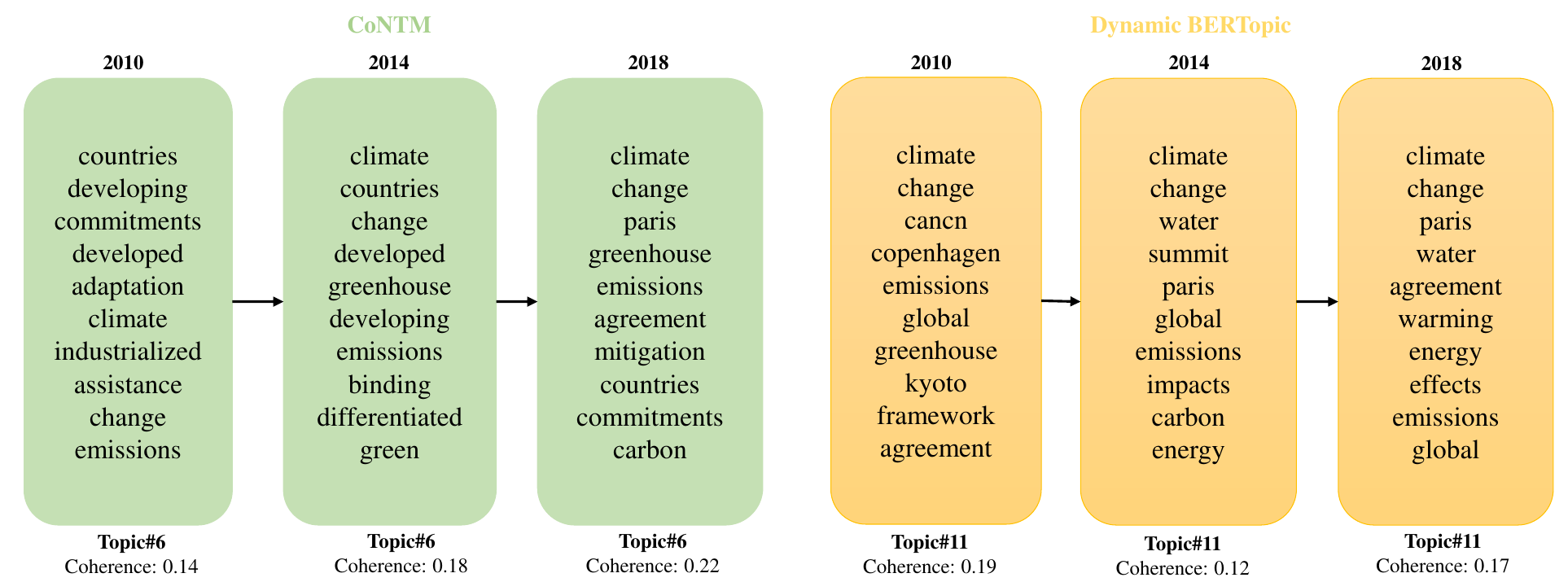} 
\caption{The figure illustrates the progression of topics\#6 (CoNTM) and topic\#11 (Dynamic BERTopics) in the UN dataset, spanning the years 2010 to 2018, both of which focus on climate change. The CoNTM model captures the emerging topic, transitioning from climate change in developed countries to the Paris Agreement which was adopted in 2015 during the UN climate change conference.}
\label{fig:un_DNTM_DBERTopics_topics}
\end{figure*}

\begin{figure*}[ht!]
\includegraphics[scale=0.45]{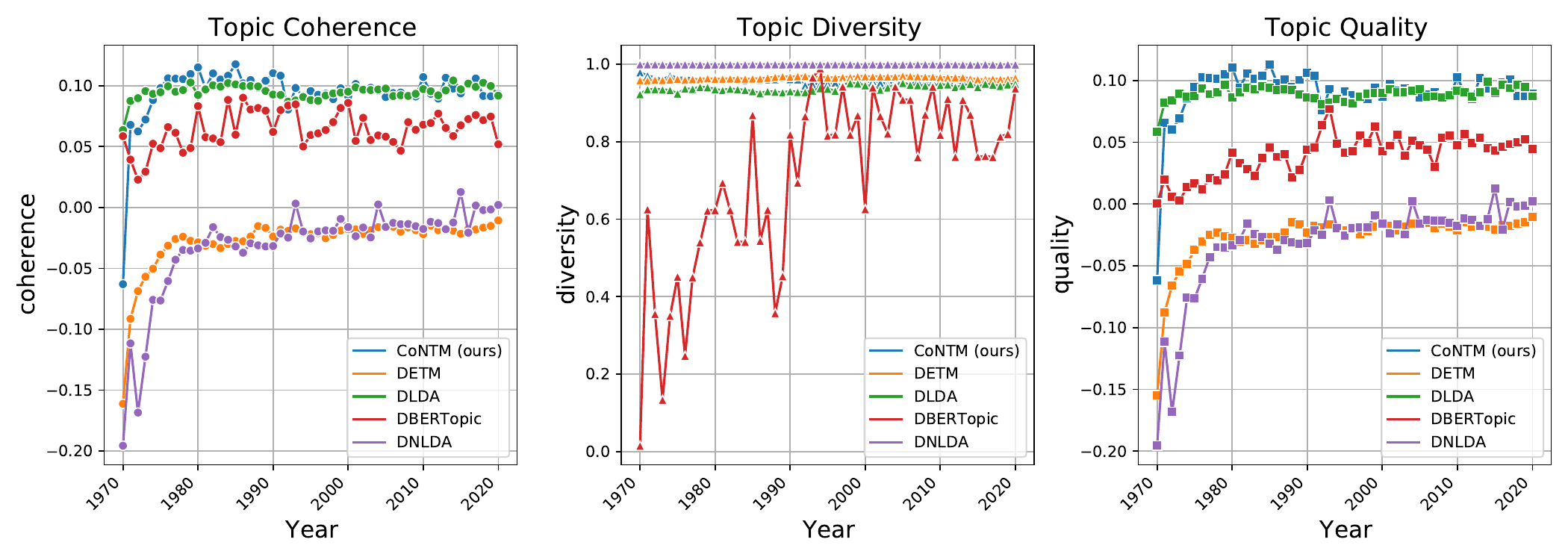} 
\caption{The figure illustrates how three measures—coherence, diversity, and topic quality—change yearly for the UN Dataset. In the case of the CoNTM model, coherence improves over time and remains stable, similar to the DLDA model. On the other hand, the topic quality of the Dynamic BERT model also increases over time, but it is lower compared to both the CoNTM and DLDA models.}\label{fig:un_tq}
\vspace{-0.5cm}
\end{figure*}

\begin{figure*}[ht!]
\includegraphics[scale=0.45]{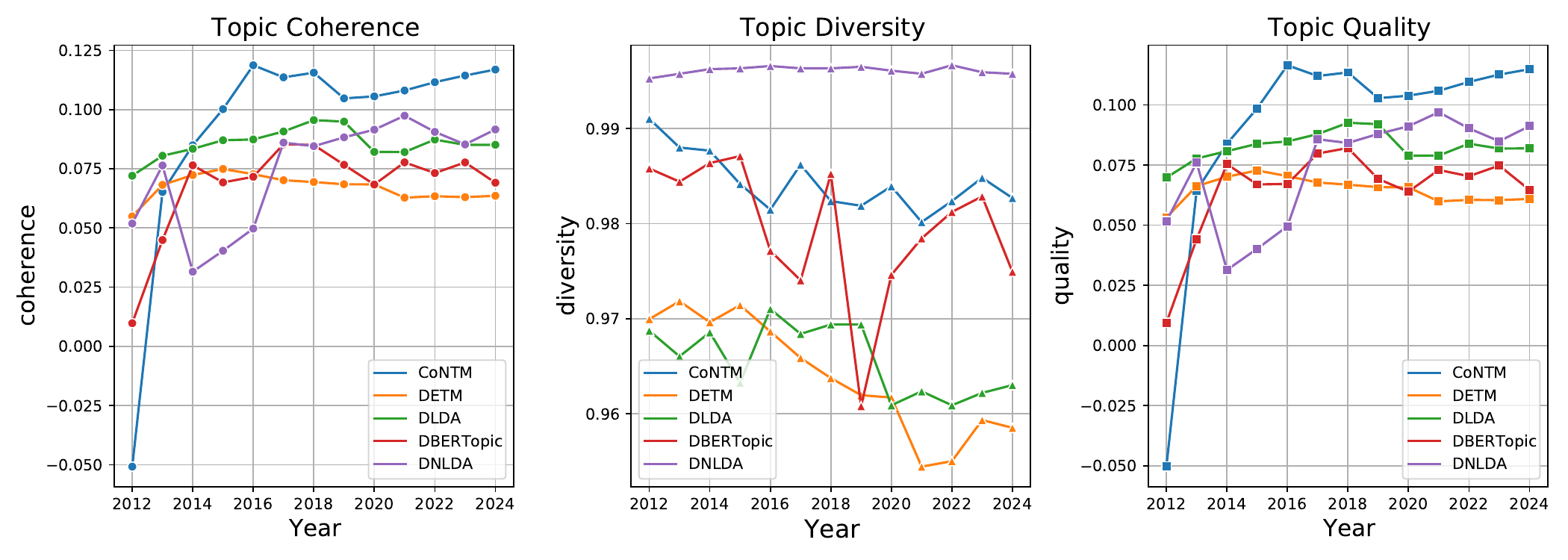} 
\caption{The figure illustrates how three measures—coherence, diversity, and topic quality—change yearly for the Arxiv Dataset. The topic quality of the CoNTM model improves over time and outperforms other models in terms of overall topic quality. }\label{fig:arxiv_tq}
\vspace{-0.5cm}
\end{figure*}

\begin{figure*}[ht!]
\includegraphics[scale=0.45]{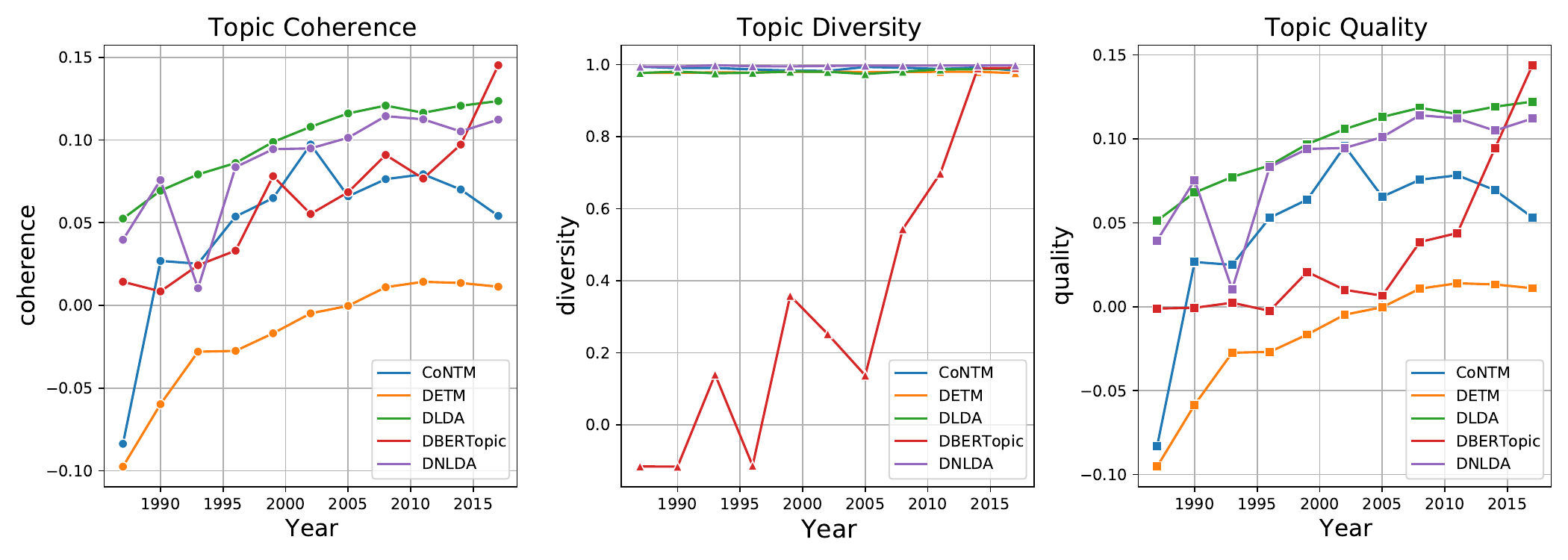} 
\caption{The figure illustrates how three measures—coherence, diversity, and topic quality—change yearly for the NIPS Dataset. The CoNTM model exhibits moderate topic quality compared to other models; this is due to insufficient documents available at the early timestamp.}\label{fig:nips_tq}
\vspace{-0.5cm}
\end{figure*}

\section{Additional Qualitative Assessment on NYT Dataset}
\label{sec: Additional Qualitative Assessment on NYT Dataset}
To compare how CoNTM, Dynamic BERTopics, DETM (Dynamic Embedded Topic Model), and DLDA (Dynamic Latent Dirichlet Allocation) capture emerging topics with good topic coherence, we should look at the clarity, continuity, and thematic relevance of the topics they produce over time. In the main paper section \ref{Qualitative Results}, the CoNTM and Dynamic BERTopics are explained. 

\textbf{Evolving Topics:} As shown in Figure \ref{fig:nyt_DETM_DLDA_topics}, DETM’s topics vary over the years, focusing on the article, king, page, and lead in 1990. The topic shifted to the election of Clinton in 1992. In DLDA, we can also observe topic changes over the years. For example, in 1990, the topics were centered around "republican", "campaign" and "political" while by 1992, the focus shifted to "clinton" "campaign", "president" and "republican". Even though both models shift the topic to the election of Clinton in 1992, the topic coherence of DLDA is better than DETM for this topic.

\textbf{Topic Coherence:} The DETM demonstrates good topic coherence, with closely related terms consistently appearing across the years. The topics reflect a stable and focused exploration of the topic politics, with less dramatic shifts compared to Dynamic BERTopics. Furthermore, the DLDA topics are less coherent, with each year building on the previous one, showing gradual changes in politics. DLDA maintains a clear thematic focus, although the evolution of topics is less coherent than in CoNTM.

\begin{figure*}[ht!]
\includegraphics[scale=0.49]{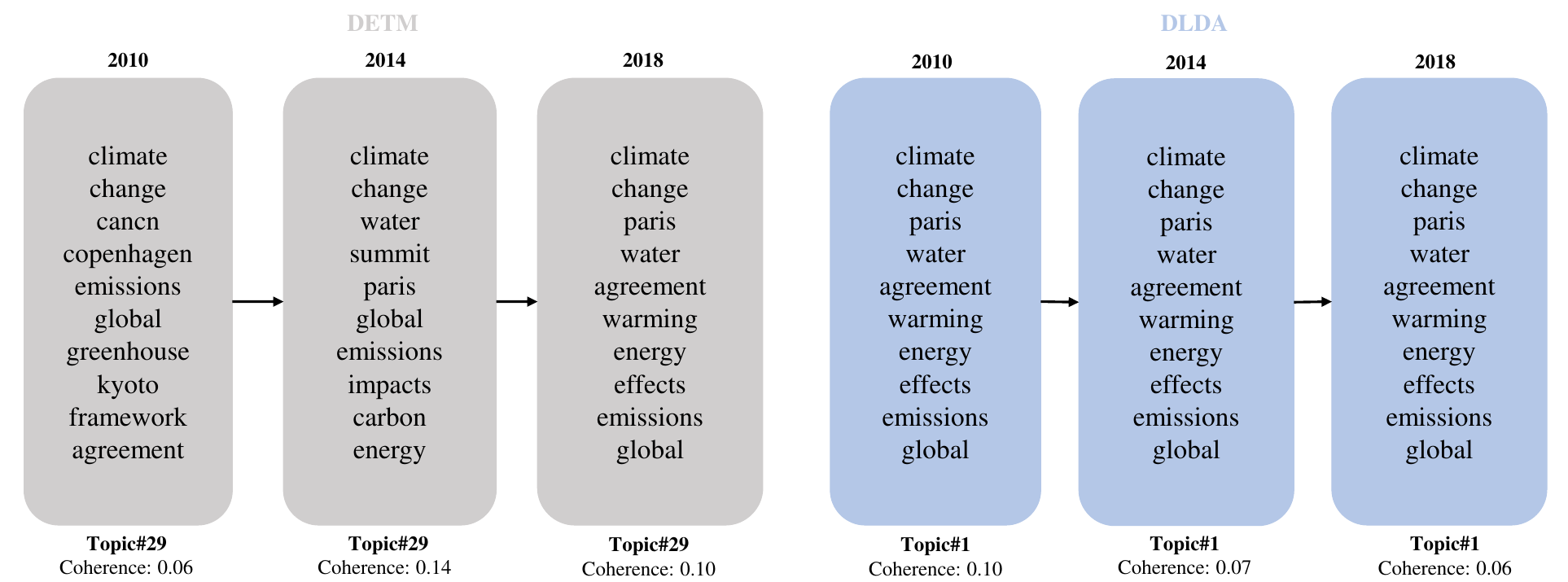} 
\caption{The figure illustrates the progression of topics\#29 (DETM) and topic\#1 (DLDA) in the UN dataset, spanning the years 2010 to 2018. Both topics focus on climate change, with the DETM model showing better topic coherence for this subject compared to the DLDA model.}
\label{fig:un_DETM_DLDA_topics}
\end{figure*}

In summary, while all models effectively capture topic evolution, CoNTM and DETM provide more coherent and gradual thematic progressions, the DLDA shows stability in topic focus, and Dynamic BERTopics excels in capturing broader shifts. CoNTM, in particular, demonstrates a strong capability for capturing emerging topics with good coherence, making it particularly useful for understanding detailed shifts in a specific thematic area over time. Example emerging topics are shown in Figure \ref{fig:NYT_emergingTopic} for the NYT dataset.

\section{Qualitative Assessment on UNDebates Dataset}
\label{sec: Qualitative Assessment on UNDebates Dataset}

To compare the CoNTM model with other models, we examine evolving topics and topic coherence using the UN dataset. Figure \ref{fig:un_DNTM_DBERTopics_topics} shows how the topic (climate change) evolves over time for the CoNTM and Dynamic BERTopic models. In both models, we can see the topic shift to the Paris Agreement in 2018, and overall topic coherence remains the same for this topic. Figure \ref{fig:UN_emergingTopic} shows three emerging topics on climate change, war, and human rights for the CoNTM model. Here, the x-axis is the timestamps, and the y-axis is the word probability. Regarding climate change, the word probability of pairs, agreement, emission, and greenhouse increased dramatically towards 2015. This is due to the Paris Agreement adopted by the UN for climate change. As shown in Figure \ref{fig:UN_emergingTopic}, the topic "war" has a high word probability for the words conflict, Iran, and Iraq, indicating a conflict in 1986.

Furthermore, Figure \ref{fig:un_DETM_DLDA_topics} illustrates the evolution of the climate change topic in the UN dataset for the DETM and DLDA models. In both models, the topic shifts to the Paris Agreement in 2018. Additionally, the DETM model's topic coherence is slightly better than the DLDA model for this topic.

\setlength{\tabcolsep}{5pt}
\begin{table*}[t!]
\begin{center} 
\begin{tabular}{l|ccc|ccc|ccc|ccc|ccc|}  
 & \multicolumn{15}{c}{\textit{20 topics}}  \\\cline{2-16}
 & \multicolumn{3}{c}{\textbf{CoNTM (ours)}} & \multicolumn{3}{|c|}{\textbf{DETM}} & \multicolumn{3}{c}{\textbf{DLDA}} & \multicolumn{3}{|c|}{\textbf{DBERTopic}} & \multicolumn{3}{|c|}{\textbf{DNLDA}}\\\hline
\textbf{Dataset} & TC & TD & TQ & TC & TD & TQ  & TC & TD & TQ &  TC & TD & TQ & TC & TD & TQ  \\ 
\hline 
NIPS & .10 & .99 & .10 &.08 & .98 & .08& .09& .96 &.08 & .06 & .48& .04& -.10& .98&-.10 \\
NYT & .19 & .99 & .19 & .12& .98 &.12 & .10 & .90 & .09 & .10 &.81 &.07 &-.08 &1.0 &-.08 \\

UN  & .12 & .98 &  .12& .09 & .97 & .09& .09 & .89 &.08 & .04 & .63&.02 & -.05& .99& -.05\\

Tweets  & -.12 & .99 & -.12 & .05& .97 &.05 & .08 & .93 & .08& .13 &.96 & .12& -.16&.99 & -.16\\

Arxiv  &.11  & .99 & .11 &.07 &.96  &.07 & .06 & .92 & .05& .05 &.94 & .04& -.07& 1.0& -.07\\
DBLP & .12 & .98 & .12 & .08& .95 &.08 & .07 & .91 & .07& .07 & .94& .07&-.06 & 1.0&-.06 \\
\hline
\end{tabular} 
\end{center}
\caption{The table compares the performance of five topic modeling algorithms: CoNTM, DETM, DLDA, Dynamic Bertopic, and DNLDA, based on topic coherence (TC), topic diversity (TD), and topic quality (TQ) with temporal reference corpus. These evaluations were conducted across six diverse datasets: NIPS, New York Times, UN, NASA Tweets, and Arxiv.}
\label{dtq_20topics}
\vspace{-0.5cm}
\end{table*}

In conclusion, the CoNTM model seems to produce more specific topics, while the Dynamic BERTopics generates broader thematic topics, indicating a difference in the level of transition each model offers. Also, the CoNTM captures the emerging topic, shifting from climate change in developed countries to the Paris Agreement in 2018, with an overall TTS score of 0.49 (see table \ref{table:tts}) for the UN dataset. In contrast, in Dynamic BERTopics, the temporal shift is more (TTS is 0.31) with a lower coherence score. The DLDA model (Figure \ref{fig:un_DETM_DLDA_topics}) shows a low coherence score for the climate change topic. Additionally, the TTS is 0.64 (see Table \ref{table:tts}), which means that the topics are changing more smoothly. A higher TTS value indicates that the topic is not evolving.

\section{Stability Across Varying Topic Size}
\label{Stability Across Varying Topic Size}
We adjusted the number of topics in the models to 20 and found that the quality of the topics generated by our model (CoNTM) exceeds that of other models across various datasets (see Table \ref{dtq_20topics}), except for the Tweets dataset. This discrepancy is likely due to the smaller number of documents in the Tweets dataset. Additionally, DBERTopic performed well because it utilizes pretrained embeddings. Notably, under the 20-topic settings, our model also outperformed other models in the NIPS dataset, which was not the case when using 50 topics.

\begin{figure*}[ht!]
\includegraphics[scale=0.58]{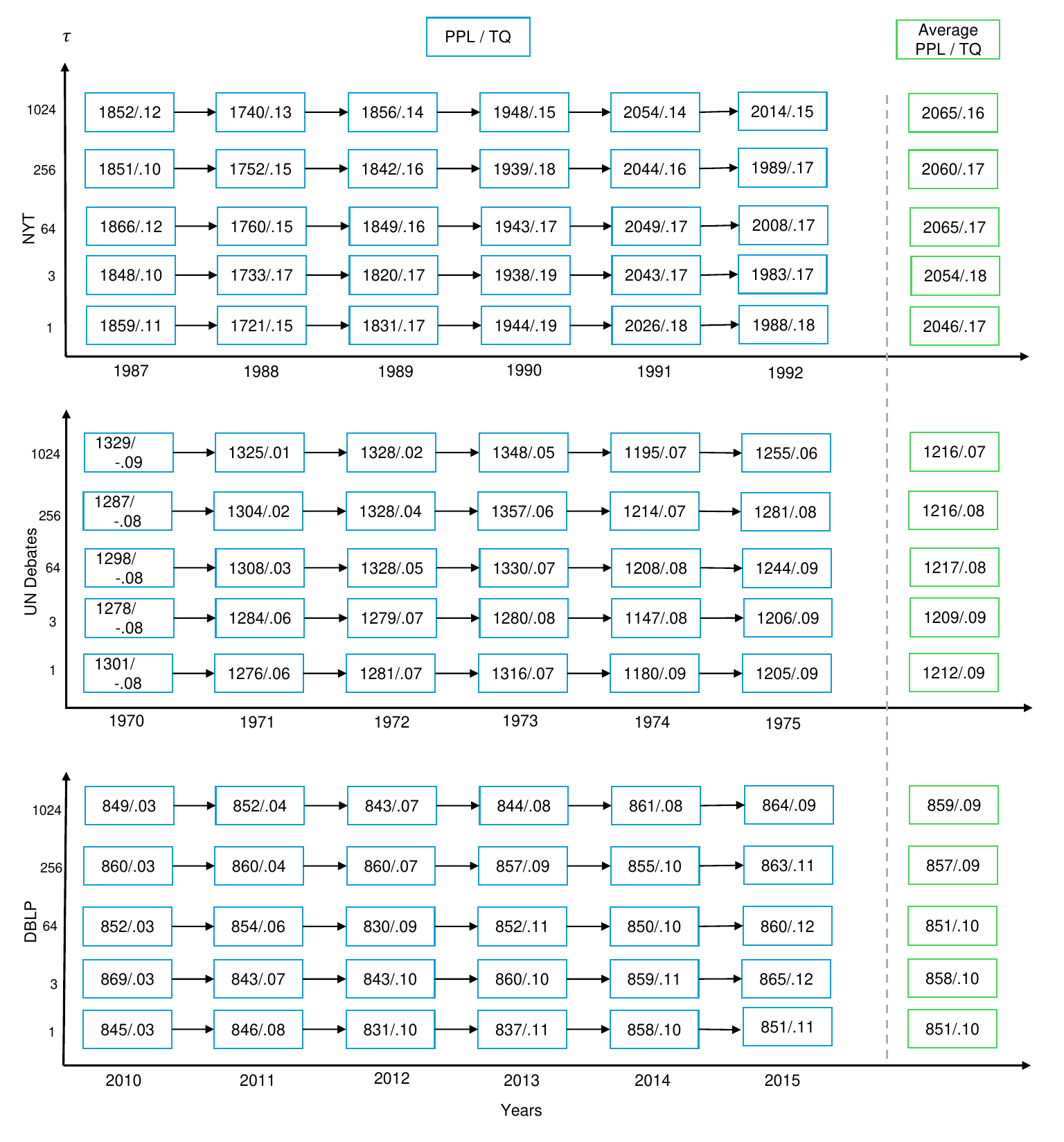} 
\caption{The figure shows the sensitivity analysis on the parameter $\rho$ ($\rho_t = \frac{1}{(\tau_0 + t)^{\kappa}}$) from Algorithm~\ref{alg:algorithm}, demonstrating a robustness to the exact choice of $\rho$. The x-axis shows perplexity/topic quality (PPL/TQ) changes over time for three different datasets such as DBLP, UnDebates, and NYT. The y-axis represents varying $\tau$ values. Here, perplexity (PPL) is the predictive perplexity on future timestamps. The blue box shows the PPL/TQ score for five timestamps, and the green box shows the average PPL/TQ for all timestamps across the respective datasets.}\label{fig:Additional ablation_rho_tau}
\vspace{-0.4cm}
\end{figure*}



For each dataset, the score for temporal topic smoothness (TTS) can be found in Table \ref{table:tts}. On average, the CoNTM model has a TTS score of 0.49, with an exception on the Tweets dataset. The Tweets have a low TTS score because the dataset lacks sufficient documents to learn more coherent and diverse topics. In summary, while the topics change gradually, the transitions are not completely smooth, allowing us to observe the evolution of topics. This is because our model learns new topics at each timestamp without forgetting previously learned information.

\begin{figure*}[ht!]
\includegraphics[scale=0.56]{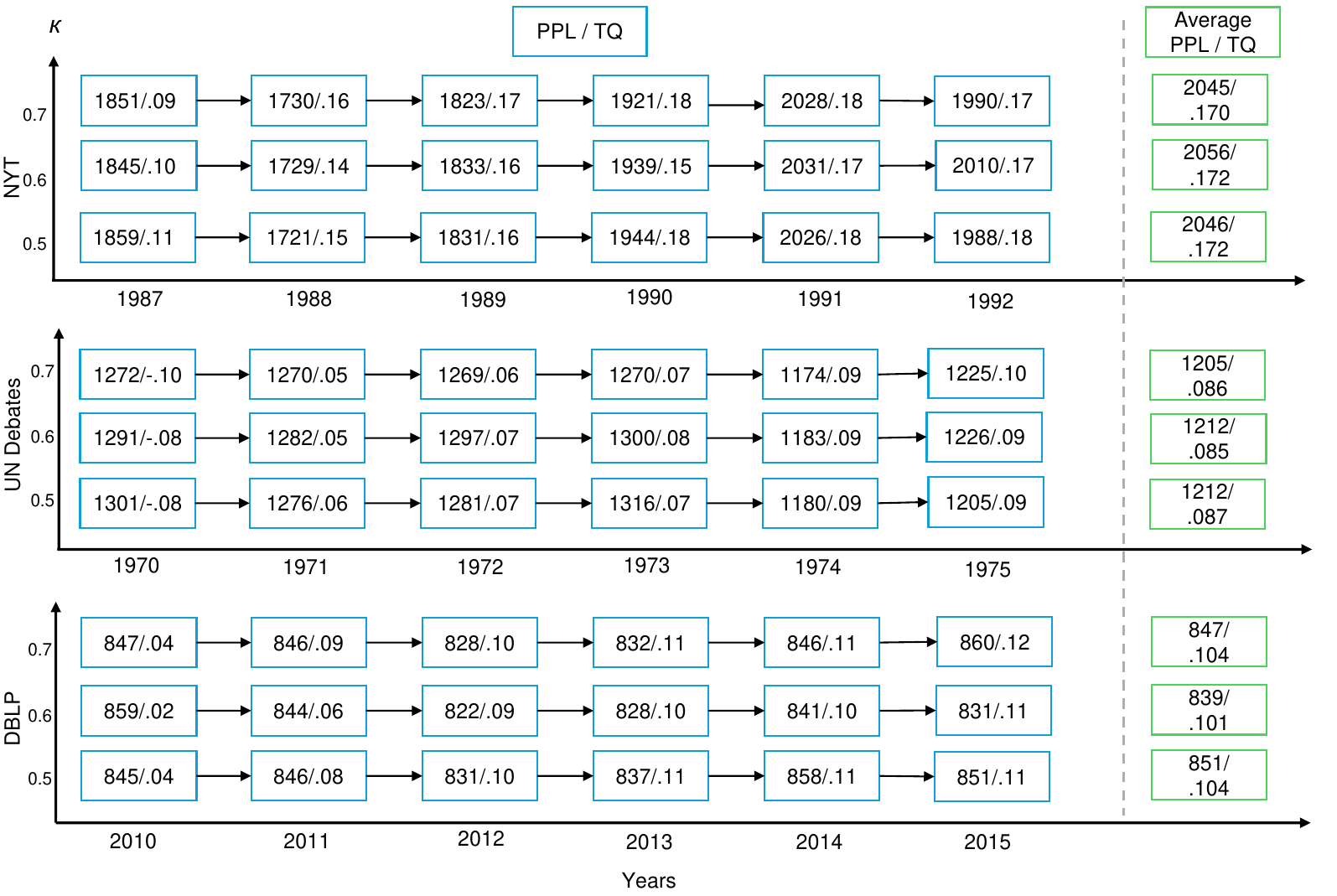} 
\caption{The figure shows the sensitivity analysis on the parameter $\rho$ ($\rho_t = \frac{1}{(\tau_0 + t)^{\kappa}}$) from Algorithm~\ref{alg:algorithm}, demonstrating a robustness to the exact choice of $\rho$. 
The blue box shows the PPL/TQ score for five timestamps, and the green box shows the average PPL/TQ for all timestamps across the respective datasets. It shows that $\rho$ does not have a significant impact on PPL/TQ.}\label{fig:ablation_rho}
\vspace{-0.3cm}
\end{figure*}

\section{Sensitivity Analysis of Rho ($\rho$)} 
\label{Sensitivity Analysis of Rho}
\begin{figure*}[ht!]
\includegraphics[scale=0.60]{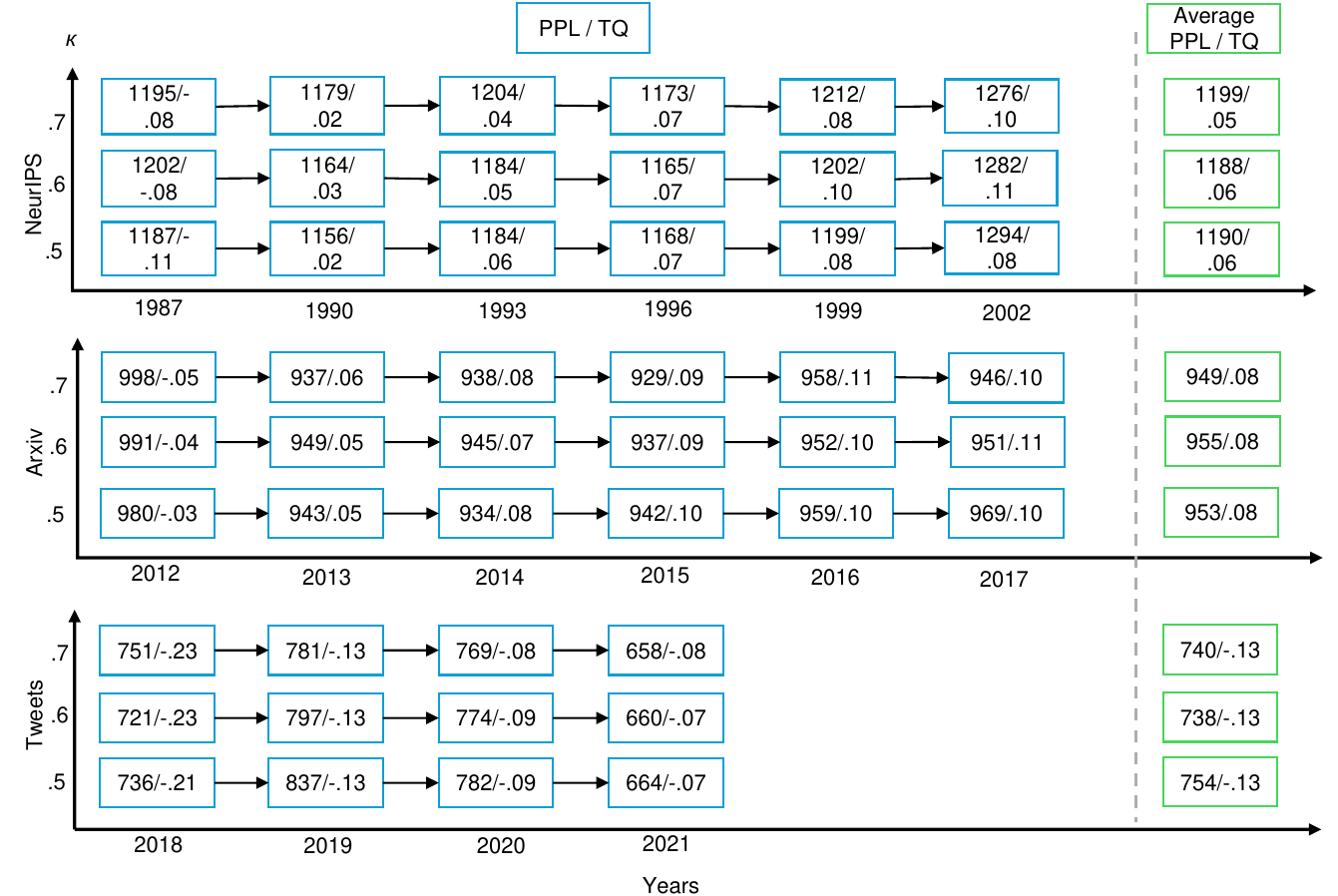} 
\caption{The figure shows the sensitivity analysis on the parameter $\rho$ ($\rho_t = \frac{1}{(\tau_0 + t)^{\kappa}}$) from Algorithm~\ref{alg:algorithm}, demonstrating a robustness to the exact choice of $\rho$. The x-axis shows perplexity/topic quality (PPL/TQ) changes over time for three different datasets such as NIPS, Arxiv, and Tweets. The y-axis represents varying $\kappa$ values. Here the perplexity (PPL), is the predictive perplexity on future timestamp. The blue box shows the PPL/TQ score for five timestamps, and the green box shows the average PPL/TQ for all timestamps across the respective datasets.}\label{fig:Additional ablation_rho_k}
\vspace{-0.4cm}
\end{figure*}

We conduct a sensitivity analysis on rho ($\rho_t = \frac{1}{(\tau_0 + t)^{\kappa}}$) from Algorithm~\ref{alg:algorithm}, step 10. Figure~\ref{fig:ablation_rho} shows that changes in the parameter $\rho$ with respect to $\kappa$ do not severely impact performance. The y-axis represents the varying $\kappa$ value, which affects the corresponding $\rho$ values. The x-axis represents different timestamps. The figure shows the effect of $\rho$ on predictive perplexity on future timestamps and topic quality over time. On average, the results indicate that $\rho$ has minimal impact on predictive perplexity or topic quality. The effect of $\kappa$ and $\tau$ for other datasets is shown in Figure \ref{fig:Additional ablation_rho_tau}, and Figure \ref{fig:Additional ablation_rho_k}, showing that a $\kappa$ value of 0.7 and a $\tau$ value of 1 are optimal.

Figure~\ref{fig:Additional ablation_rho_tau} illustrates that variations in $\rho$ with respect to $\tau$ have a minimal impact on performance. The y-axis represents the varying $\tau$ value, which influences the corresponding $\rho$ values. The x-axis represents different timestamps. The figure shows the effect of $\tau$ on predictive perplexity at future timestamps and topic quality over time. The results indicate that a $\tau$ value of 1 yields lower predictive perplexity and higher topic quality for the CoNTM model.

Figure \ref{fig:Additional ablation_rho_k} shows the sensitivity analysis of the parameter $\rho$ with respect to $\kappa$ for the NIPS, Arxiv, and Tweets datasets. The results indicate that $\kappa$ does not significantly impact topic quality and predictive perplexity. However, on average, a $\kappa$ value of 0.7 yields better results.

\section{Statistical Significance Testing}
\label{Statistical Significance Testing}
The t-test results indicate significant differences in perplexity values for some datasets, while others do not show substantial differences. For datasets such as NIPS, NYT, and DBLP, the p-values are well below 0.05, indicating a statistically significant difference between the models. However, for datasets like un and tweets, the p-values are above 0.05, suggesting no strong evidence for a difference.

\setlength{\tabcolsep}{1pt}
\renewcommand{\arraystretch}{1.2} 
\begin{table}[ht!]
\centering
\begin{tabular}{l|cc}
\hline
\textbf{Dataset} & \textbf{t-statistic} & \textbf{p-value} \\
\hline
NIPS  & -8.716 & 6.365591e-08  \\
NYT & -21.555 & 1.828662e-13 \\
UN &  -4.014 & 5.186161e-04  \\
Tweets & -0.928 & 3.594985e-01 \\
Arxiv & 0.829 & 4.128483e-01 \\
DBLP &4.170 &1.955962e-04
 \\
\hline
\end{tabular}
\caption{Statistical Significance Test (t-test) on predictive perplexity between CoNTM and DETM with twenty different seed values.}
\label{table:t-test}
\vspace{-0.6cm}
\end{table}

\begin{figure*}[ht!]
\begin{center} 
\includegraphics[scale=0.32]{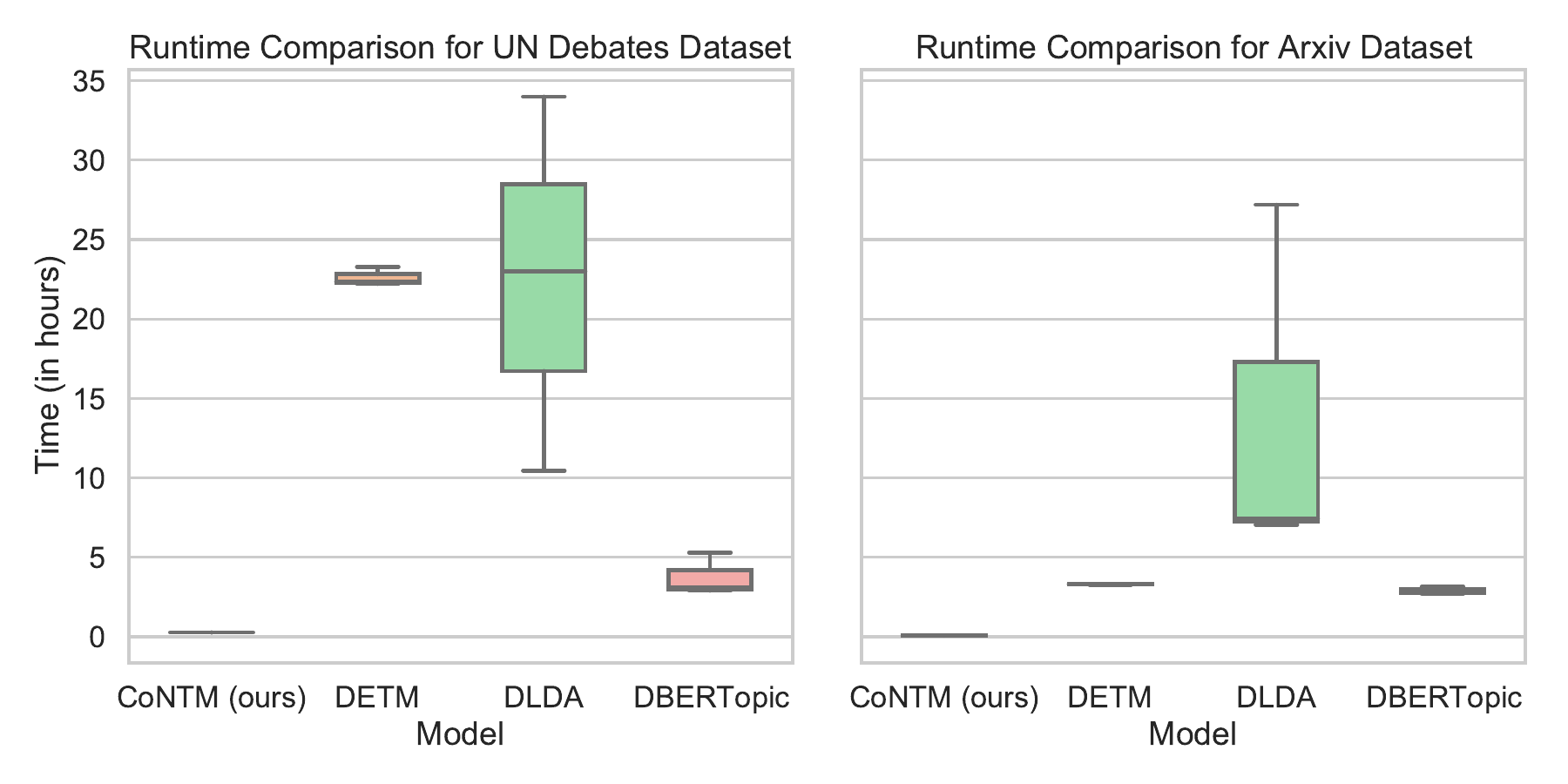} 
\caption{Runtime comparison (in hours) for evolving topics on UN and the Arxiv dataset shows that the CoNTM model has the shortest runtime, outperforming all other tested models.}\label{fig:runetime}
\end{center} 
\end{figure*}

\begin{figure*}[ht!]
\includegraphics[scale=0.31]{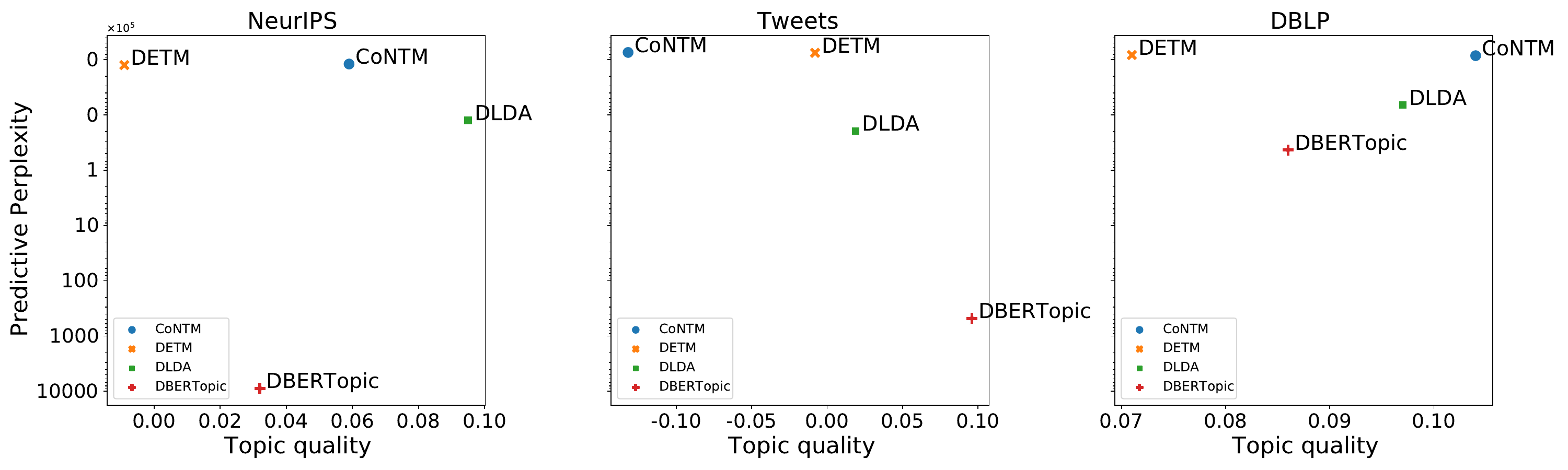} 
\caption{The figure shows the model performance quantitatively for NIPS, Tweets, and DBLP datasets. In the plot, the top-right corner indicates that the model achieves high topic quality and low predictive perplexity. Our model (CoNTM) outperformed the other models for DBLP datasets in terms of both quality and perplexity.}\label{fig:quality_vs_PredictivePerplexity_tweets_nips_dblp}
\vspace{-0.4cm}
\end{figure*}

\section{Runtime Comparison}

A comparative analysis of their runtimes is essential in evaluating the efficiency of various topic modeling algorithms (see Figure \ref{fig:runetime}). This section discusses the runtime performance of four different models, CoNTM (our proposed model), DETM, DLDA, and DBERTopic, across two datasets: the UN Dataset and the Arxiv Dataset.

For the UN Dataset, the runtime analysis reveals a significant variance in the computational efficiency of the models. The CoNTM model shows a remarkable performance advantage, with the lowest runtime among the models tested. Following CoNTM, DETM and DLDA exhibit moderately higher runtimes.

The runtime comparison maintains a similar trend in the context of the Arxiv dataset. The CoNTM model again stands out for its efficiency, underlining the effectiveness of our optimization strategies in reducing computational overhead. The other models, DBERTopic, DETM, and DLDA, follow in increasing order of runtime, consistent with the findings from the UN Datasets.

\section{Topic Coherence vs Diversity}
\label{Topic Coherence vs Diversity}

\begin{figure*}[h!]
\includegraphics[scale=0.31]{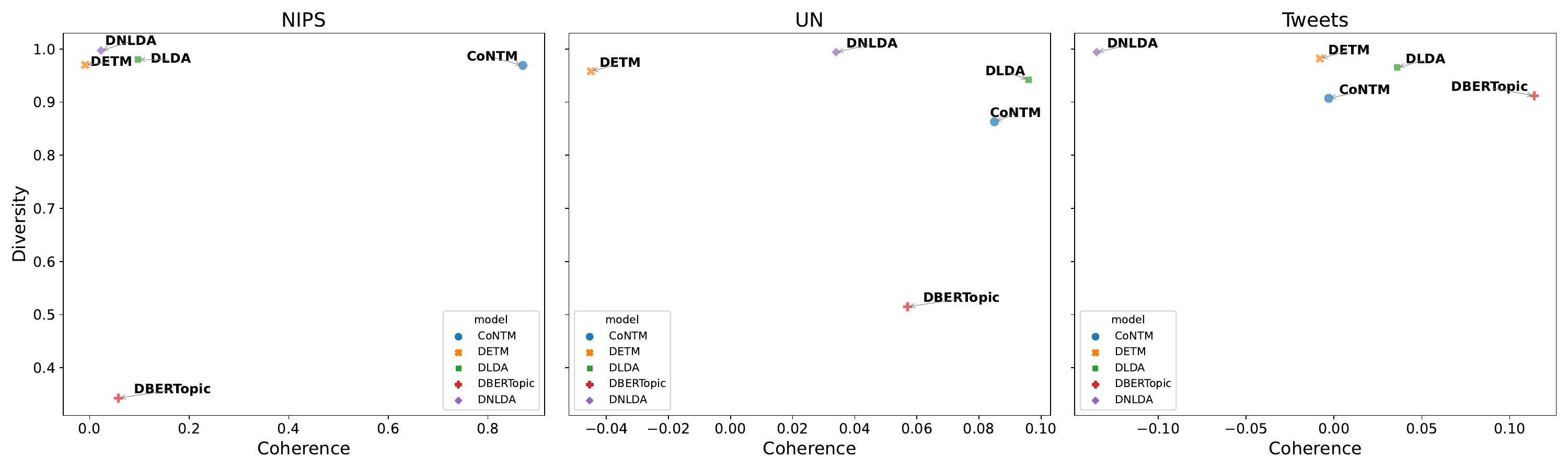} 
\caption{The figure shows the model performance quantitatively for the NIPS, UN, and Tweets datasets. The top-right corner indicates that the model achieves high topic quality and low predictive perplexity.}\label{fig:all_coherence_vs_diversity_NIPS_UN_Tweets}
\vspace{-0.1cm}
\end{figure*}

The quantitative analysis of additional datasets such as NIPS, UN, and Tweets is shown in Figure \ref{fig:all_coherence_vs_diversity_NIPS_UN_Tweets}. On the Tweets dataset, DBERTopic demonstrates notably high topic quality. This can be due to its use of pre-trained word embeddings, which significantly enhance topic coherence, especially in smaller datasets. The use of pre-trained embeddings allows DBERTopic to capture semantic relationships more effectively, leading to improved performance in terms of topic quality and coherence compared to other models on limited data.

\section{Additional Quantitative Results}
\label{Additional Quantitative Results}
This section analyzes CoNTM and compares it with other models on the NIPS, Tweets, and DBLP datasets. 

\textbf{Topic Quality vs Predictive Perplexity:} Our evaluation reveals that CoNTM achieves superior topic quality while maintaining lower predictive perplexity on the DBLP dataset, as illustrated in Figure  \ref{fig:quality_vs_PredictivePerplexity_tweets_nips_dblp}. For the NIPS dataset, although the CoNTM topic quality is slightly lower than that of the DLDA model, it consistently outperforms all other models in predictive perplexity. This demonstrates CoNTM’s ability to generate coherent topics while maintaining strong predictive performance. The y-axis is presented on a logarithmic scale, giving better visibility into the smaller values without losing the large one.

\section{Emerging Topics}
\label{Emerging Topics}

\begin{figure}[htp]
    \centering
    \begin{minipage}[b]{0.45\textwidth}
        \centering
        \includegraphics[width=\textwidth]{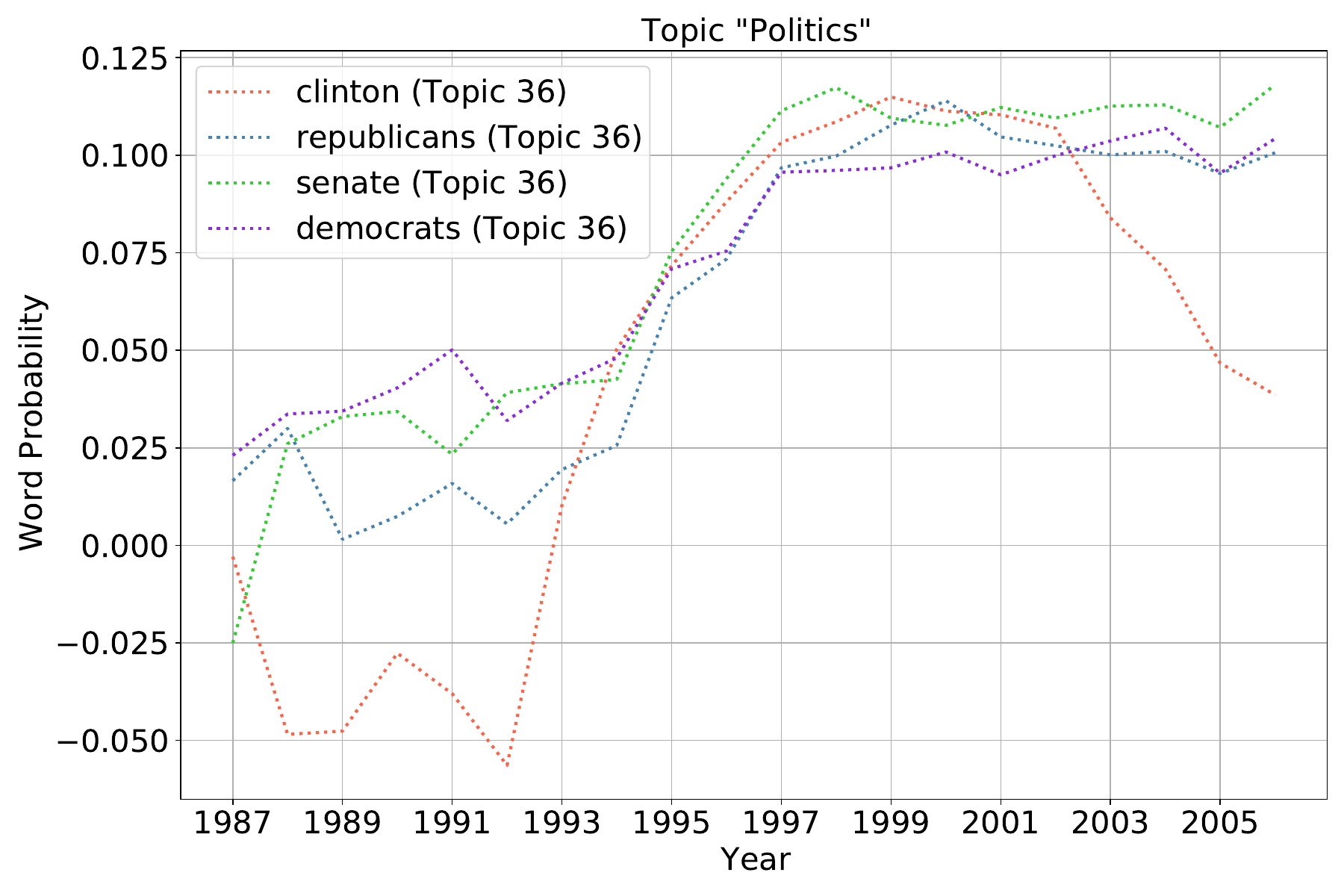}
        \caption{The figure shows emerging topics "Politics" in the CoNTM (our) model for the NYT dataset. The y-axis shows the word probability of topic words from the trained model and the x-axis shows the timestamp.}
         \label{fig:NYT_emergingTopic}
    \end{minipage}
\end{figure}

This section provides a few emerging topics in the NYT and UN dataset. Figure \ref{fig:NYT_emergingTopic} shows the emerging topic "politics," with word probability on the y-axis and timestamps on the x-axis for the NYT dataset. In 1997, \textit{Clinton} was inaugurated for his second term as the 42nd President of the U.S., with his final years in office spanning from 1999 to 2000. Following this period, there is a noticeable decline in the word probability of \textit{Clinton}.

\begin{figure}[htp]
    \centering
    \begin{minipage}[b]{0.45\textwidth}
        \centering
        \includegraphics[width=\textwidth]{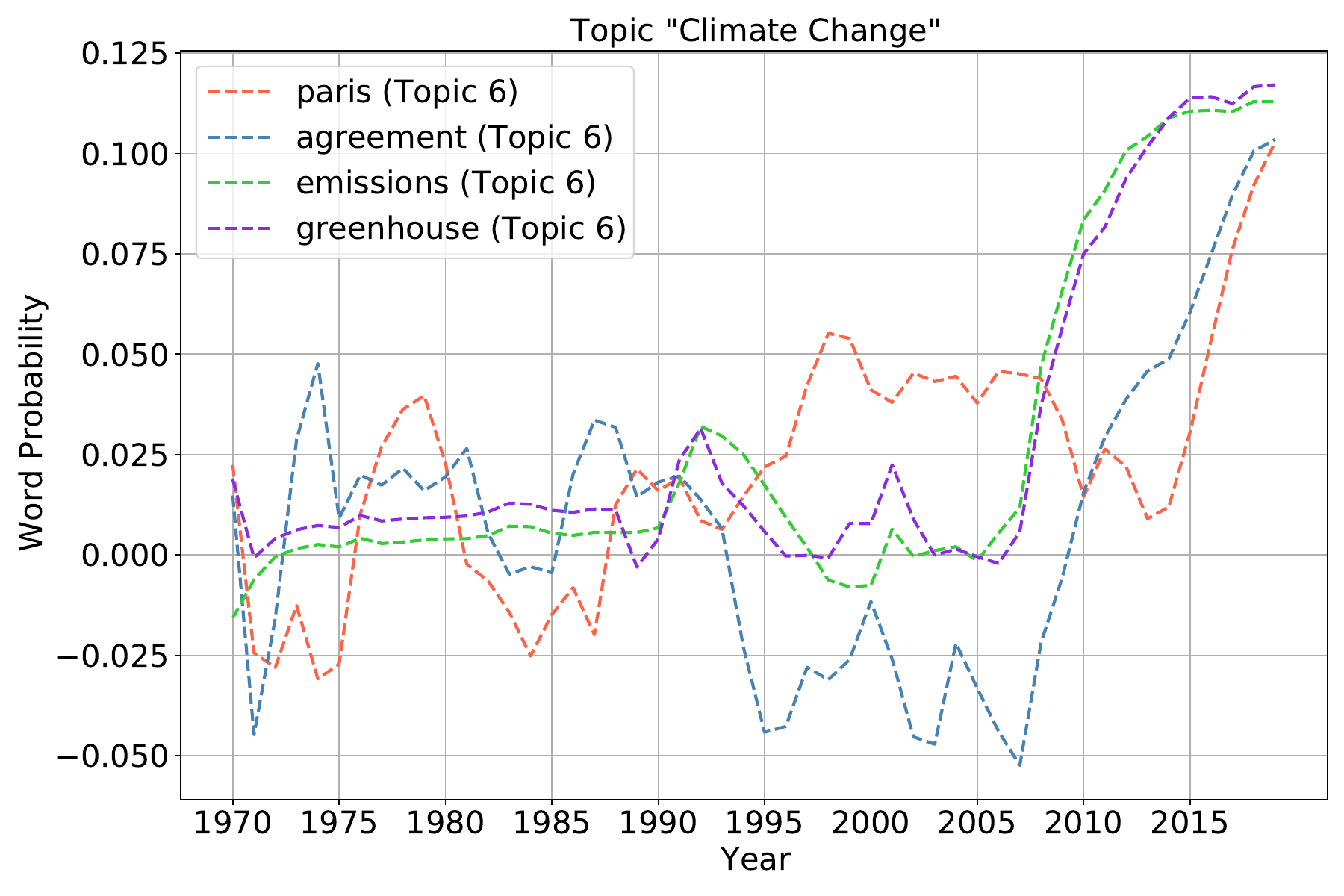}
    \end{minipage}
    \hfill
    \begin{minipage}[b]{0.45\textwidth}
        \centering
        \includegraphics[width=\textwidth]{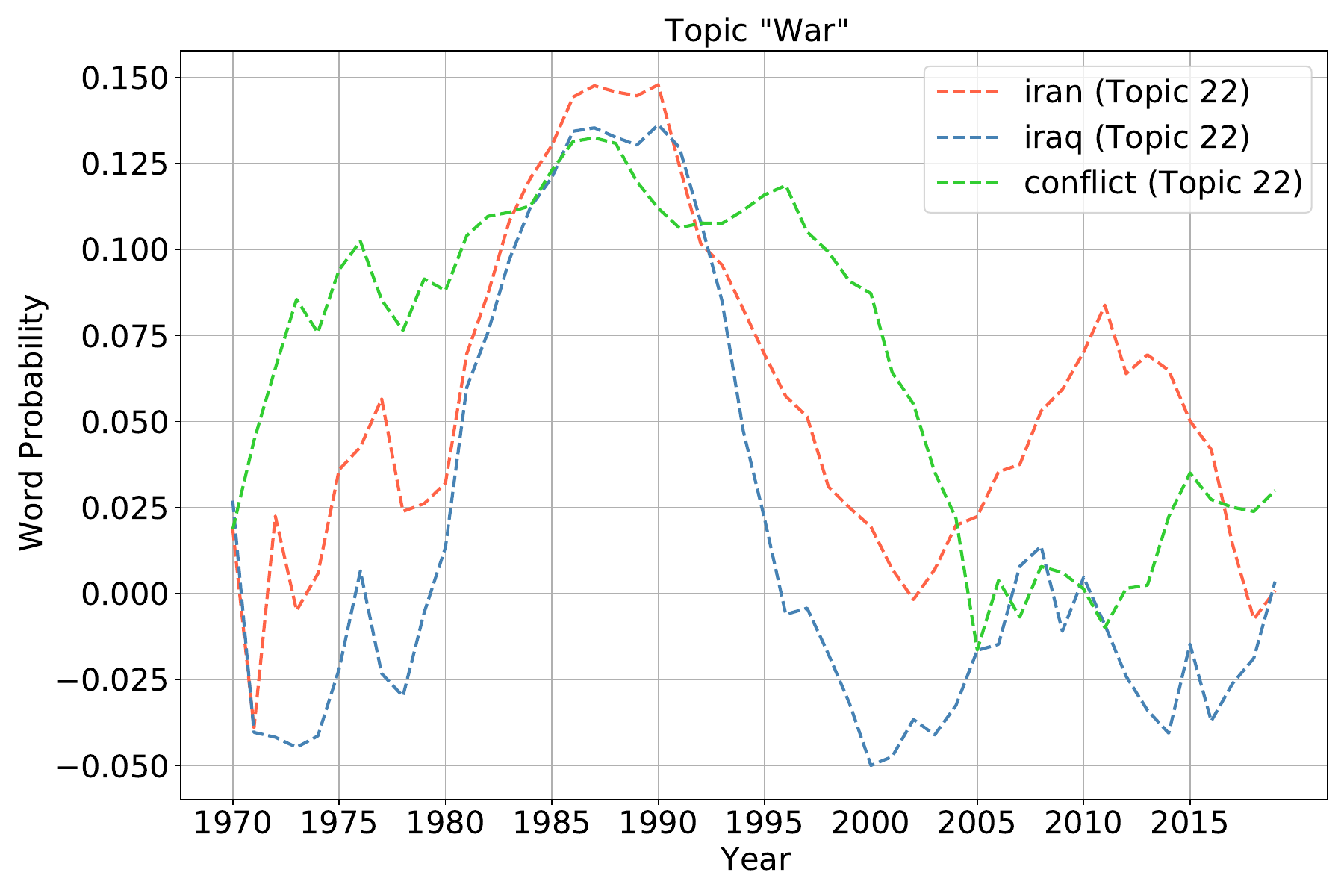}
    \end{minipage}
    \hfill
    \begin{minipage}[b]{0.45\textwidth}
        \centering
        \includegraphics[width=\textwidth]{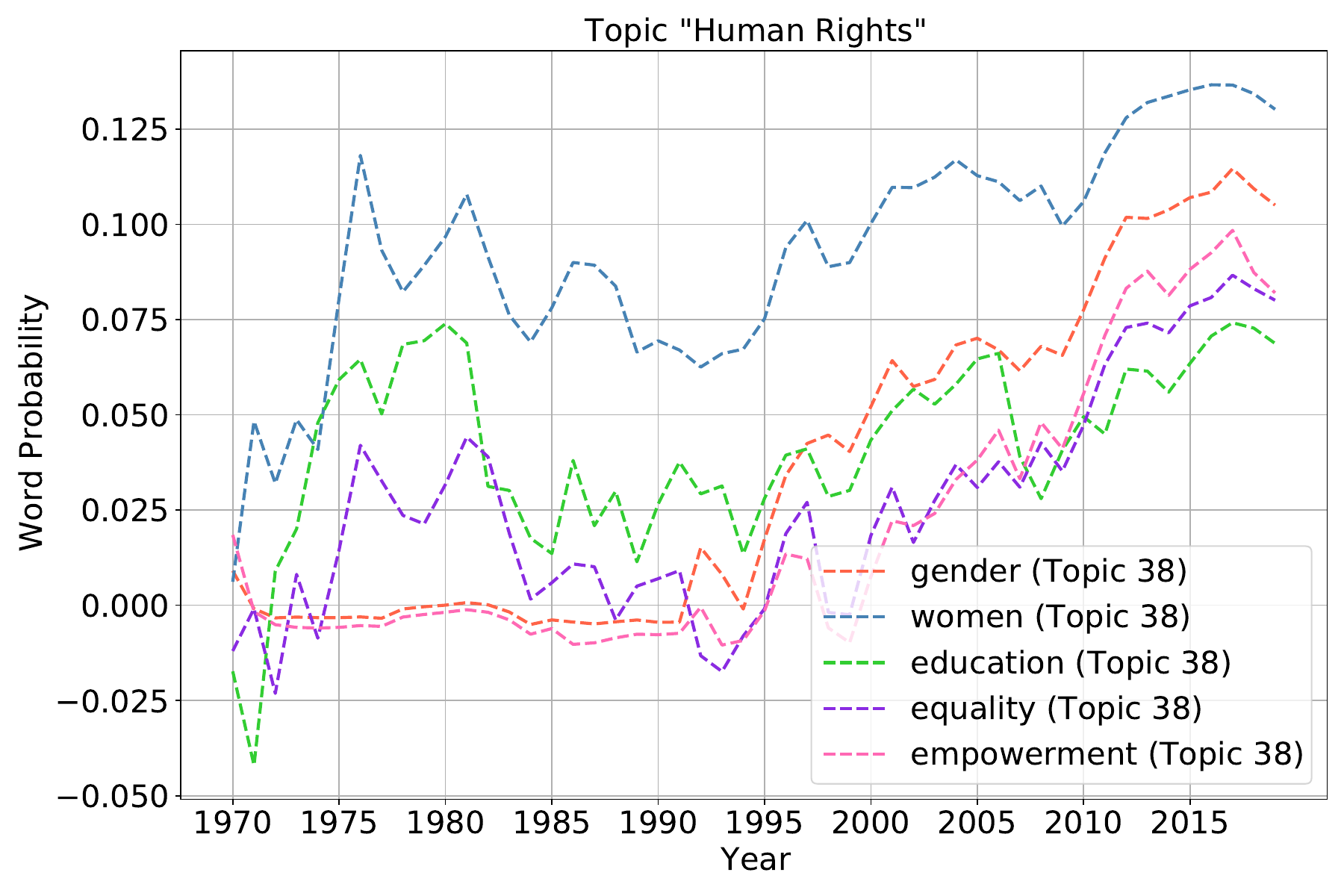}
        \caption{The figure shows three emerging topics "Climate Change", "War", and "Human Rights" in the CoNTM (our) model for the UN dataset. The y-axis shows the word probability of topic words from the trained model and the x-axis shows the timestamp.}
        \label{fig:UN_emergingTopic}
    \end{minipage}
    
\end{figure}

Figure \ref{fig:UN_emergingTopic} shows the emerging topics "Climate Change", "War", and "Human Rights" from the UN dataset. For the "Climate Change" topic, the word probability of Paris, agreement, emissions, and greenhouse is notably high, aligning with the negotiations and implementation of the Paris Agreement in 2015.

\section{Generalization via Predictive Perplexity (PPL)}
\label{PPL}
\begin{figure*}[htp]
    \centering
    \begin{minipage}[b]{0.70\textwidth}
        \centering
        \includegraphics[width=\textwidth]{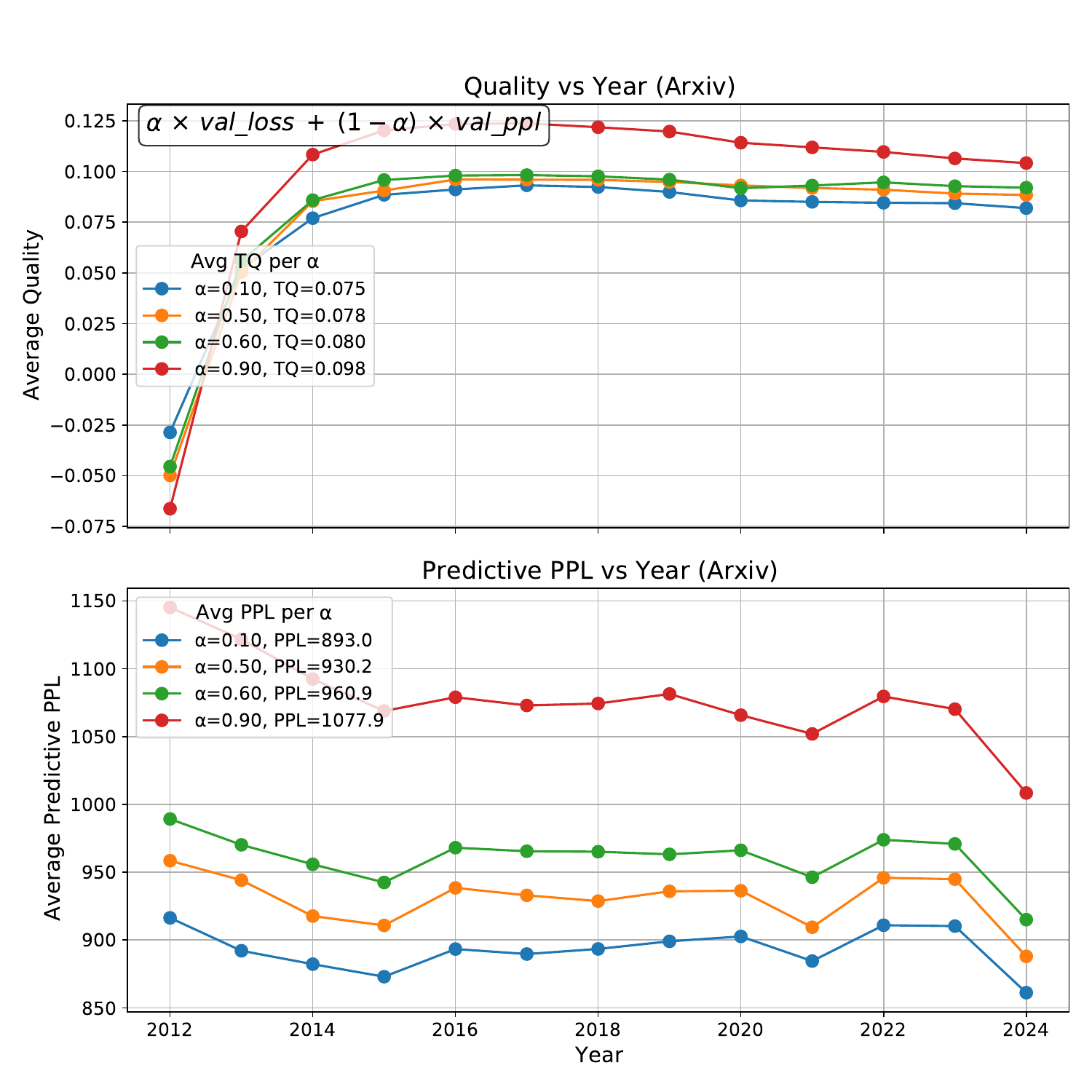}
        \caption{The figure illustrates the trade-off between predictive perplexity and topic quality for Arxiv dataset.}
        \label{fig:tradeoff_Arxiv}
    \end{minipage}
    \hfill
    \begin{minipage}[b]{0.70\textwidth}
        \centering
        \includegraphics[width=\textwidth]{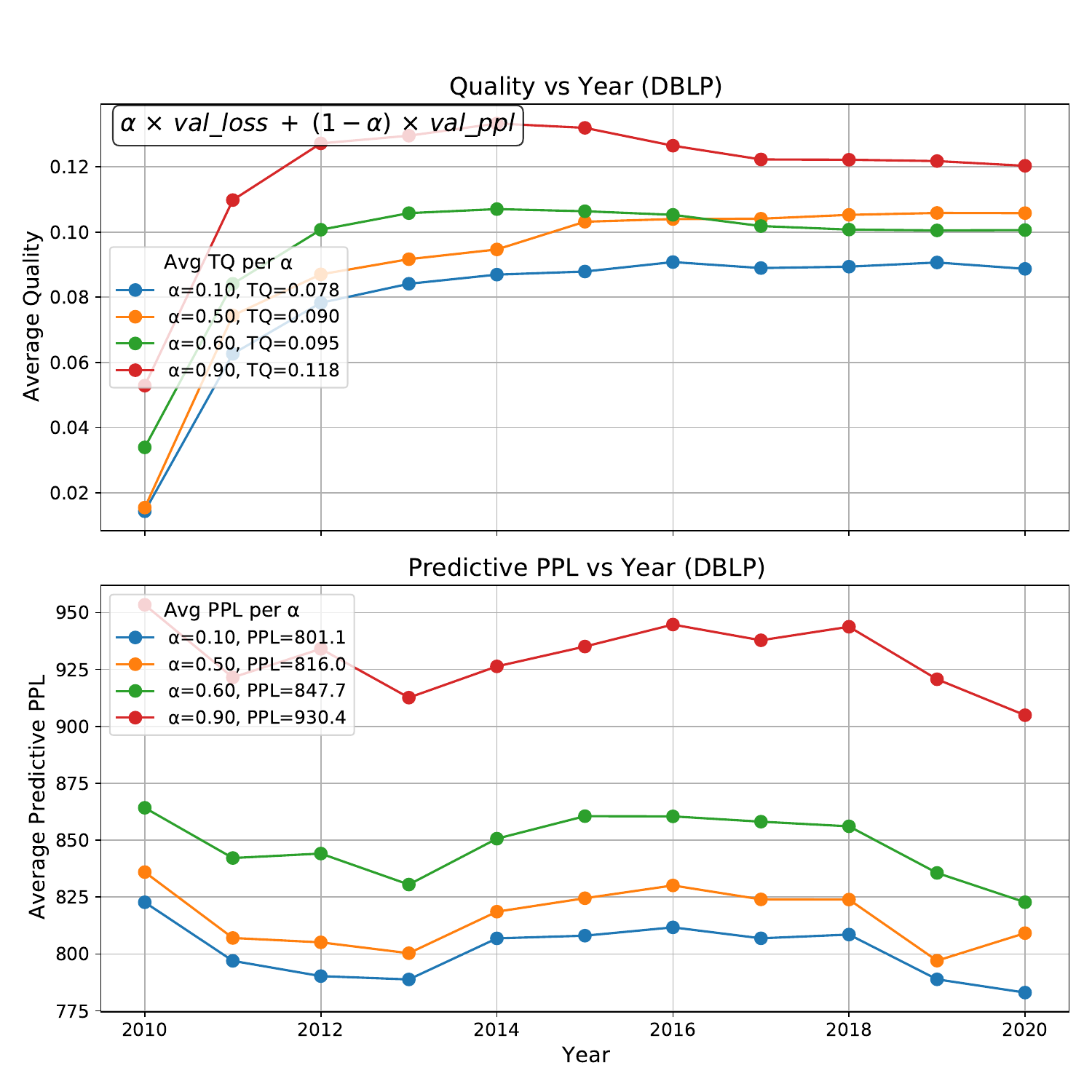}
        \caption{The figure illustrates the trade-off between predictive perplexity and topic quality for DBLP dataset.}
        \label{fig:tradeoff_NYT}
    \end{minipage}
    
\end{figure*}

Perplexity is a standard metric for evaluating probabilistic language models. Table \ref{table:ppl} shows predictive perplexity on future timestamp data for all models, where we compute the perplexity of documents in timestamp $(t+1)$ based on the model trained at timestamp $(t)$. The CoNTM model achieves the lowest predictive perplexity (lower is better) in all datasets except Tweets, demonstrating its ability to predict unseen data. The Tweets are very short, providing limited context for the model to capture meaningful patterns, leading to a decrease in performance compared to longer documents.  In conclusion, CoNTM demonstrates good performance, while DLDA and Dynamic BERTopic show significantly lower performance. The t-test for CoNTM and DETM is detailed in Appendix \ref{Statistical Significance Testing}. Also, the value presented in the table is the average of three random seeds.  The DNLDA perplexity score is not included in the table due to its poor performance. Figures \ref{fig:tradeoff_Arxiv}, and \ref{fig:tradeoff_NYT} show the trade-off between predictive perplexity and topic quality on the Arxiv and DBLP datasets. The results shown are based on three randomly selected seeds. In the figure, when $\alpha = 0.90$, topic quality (where higher is better) is high, but predictive perplexity (where lower is better) is also high. Conversely, when $\alpha = 0.10$, topic quality decreases, but predictive perplexity improves. This illustrates a trade-off between topic quality and predictive perplexity.

\setlength{\tabcolsep}{1pt}
\renewcommand{\arraystretch}{1.0} 
\begin{table}[ht]
\centering
\begin{tabular}{l|cccc}
\hline
\textbf{Dataset} & \textbf{CoNTM} & \textbf{DETM} & \textbf{DLDA} & \textbf{DBERTopic}\\
\hline
NIPS  & \textbf{1199} & 1254 &12.5K & 8.9E8 \\
NYT & \textbf{2045} & 2172& 85.9K & 337K \\
UN & \textbf{1205} &1268 &10.9K  & 1.1E8 \\
Tweets & \textbf{740} &752  & 19.5K & 4.8E7\\
Arxiv & \textbf{949} & 953 & 7921 & 99K\\
DBLP & 847 & \textbf{823} & 6634 & 43K\\
\hline
\end{tabular}
\caption{The table shows the average predictive perplexity over three runs for six diverse datasets. The CoNTM model provides better predictive perplexity on almost all datasets.}
\label{table:ppl}
\vspace{-0.8cm}
\end{table}

\end{document}